\documentclass[twoside]{article}

\usepackage[accepted]{aistats2025}
\usepackage[utf8]{inputenc}
\usepackage[T1]{fontenc}
\usepackage{hyperref}
\usepackage{url}
\usepackage{color}
\usepackage{xcolor}

\usepackage{qiangstyle}
\usepackage{amsthm}
\usepackage{amsmath}
\usepackage{amssymb,bbm}
\usepackage{amsfonts}
\usepackage{comment}
\usepackage{cancel}
\usepackage{nicefrac}
\usepackage{mathtools}
\usepackage{algorithm}
\usepackage{algorithmic}
\newtheorem{proposition}{Proposition}
\usepackage{natbib}

\usepackage{multirow}
\usepackage{tabularx}
\usepackage{adjustbox}
\usepackage{diagbox}
\usepackage{booktabs}
\usepackage{float}
\usepackage{graphicx}
\usepackage{wrapfig}
\usepackage{caption,subcaption}
\usepackage{epsf}
\usepackage{psfrag}
\usepackage{tikz}

\usepackage{microtype}
\usepackage{fancyhdr}
\usepackage{enumitem}
\usepackage{pbox}
\usepackage{titlesec}
\titlelabel{\thetitle.\quad}

\newcommand{\ours}{\textit{H-Fac}}

\begin{document}

%

%

\twocolumn[

\aistatstitle{Memory-Efficient Optimization with Factorized Hamiltonian Descent}

\aistatsauthor{Son Nguyen \hspace{1.8cm} Lizhang Chen \hspace{1.8cm}  Bo Liu \hspace{1.8cm} Qiang Liu}
\vspace{0.08in}
\aistatsaddress{ The University of Texas at Austin \\
\texttt{\{sonnv77,lzchen,bliu,lqiang\}@utexas.edu} } ]

\begin{abstract}
Modern deep learning heavily depends on adaptive optimizers such as Adam and its variants, which are renowned for their capacity to handle model scaling and streamline hyperparameter tuning. However, these algorithms typically experience high memory overhead caused by the accumulation of optimization states, leading to a critical challenge in training large-scale network models. In this study, we introduce a novel adaptive optimizer, \ours{}, which incorporates a memory-efficient factorization approach to address this challenge. By employing a rank-1 parameterization for both momentum and scaling parameter estimators, \ours{} reduces memory costs to a sublinear level while maintaining competitive performance across a wide range of architectures. We develop our algorithms based on principles derived from Hamiltonian dynamics, providing robust theoretical underpinnings in optimization dynamics and convergence guarantees. These optimization algorithms are designed to be both straightforward and adaptable, facilitating easy implementation in diverse settings. 
\end{abstract}

\section{INTRODUCTION}
Optimization algorithms play an indisputable role in the remarkable development of AI, especially in the realm of modern deep learning. In recent years, the emergence of breakthroughs in architectural innovation~\citep{bommasani2021opportunities} and their practical applications, has further promoted the necessity for embracing efficient training paradigms, which encompass optimization algorithms striking a balance between performance and manageable memory costs. 

Stochastic gradient descent (SGD) is widely regarded as the standard algorithm for training deep learning 
\enlargethispage{\baselineskip}
models, supported by extensive theoretical foundations~\citep{csimcsekli2019heavy, smith2021origin, tian2023recent, zhou2020towards}. However, it requires thorough tuning of hyperparameters and frequently exhibits undesirable convergence rates when applied to many contemporary architectures~\citep{devlin2018bert, vaswani2017attention, zhang2020adaptive}. Meanwhile, adaptive moment-based methods such as Adam~\citep{kingma2014adam}, AdaGrad~\citep{duchi2011adaptive}, AMSGrad~\citep{reddi2019convergence}, and variants, can adjust the learning rate for each parameter throughout the optimization process by utilizing a cumulative second moment of the gradient statistics. Although the theoretical aspects have not yet been fully exploited, these methods show empirical performance that surpasses SGD across multiple domains and often provides better convergence properties in practice~\citep{zhang2020adaptive, loshchilov2017decoupled}, making them highly appealing for large-scale applications. However, maintaining momentum and per-coordinate scaling parameter accumulators in these algorithms will significantly increase memory overhead. This barrier typically restricts model size, reduces the number of examples per mini-batch, or limits communication in decentralized training~\citep{konevcny2016federated, li2020federated, liu2024communication}, thereby negatively affecting convergence and accuracy. 

Several memory-efficient optimizers have been devised to address this problem. In particular, Adafactor~\citep{shazeer2018adafactor} is a highly efficient algorithm that was designed for low memory usage and greater scalability. It achieves substantial memory savings by employing a non-negative matrix factorization approach to decompose the second-moment accumulator into two distinct rank-1 factors. This technique not only minimizes memory footprint but also enhances computational efficiency. However, to avoid performance degradation, Adafactor still needs to maintain the first-moment estimator, particularly when applied to large-scale models. This could potentially limit its applicability in memory-constrained environments. Additionally, Adafactor does not yet have a solid theoretical foundation, leaving its underlying methodologies and effectiveness without formal justification.

In this work, we propose to approach the aforementioned challenges through the powerful tool of Hamiltonian dynamics. Motivated by some inspirational works \citep{maddison2018hamiltonian, chen2023lion}, we explore how Hamiltonian principles in modeling dynamic systems can inform and improve optimization strategies, especially for memory efficiency purposes. Our contributions can be summarized in the following:
\begin{itemize}
    \item We demonstrate that Adafactor can be deduced from an ordinary differential equation (ODE) that solves a minimization problem with constrained factors. This novel perspective allows us to derive other algorithms with the same principles.
    \item We propose a new class of efficient optimization algorithms that embraces both momentum-based and adaptive methods. By employing rank-1 parameterization for momentum and scaling parameter, our optimizers can offer sublinear memory costs, comparable to that of vanilla SGD without momentum. To the best of our knowledge, our proposal is the first endeavor to exploit Hamiltonian dynamics in developing a class of memory-efficient optimization algorithms that utilize factorized gradient statistic estimators.
    \item Distinct from existing optimization techniques for memory efficiency, our algorithm can offer clear insights into optimization dynamics and convergence guarantees, which are naturally inherited from the fundamental theory of Hamiltonian mechanics. This guarantee is significant because it can be broadly applied to a wide range of proposed optimizers, without relying on strong assumptions about underlying models or objective functions.
    \item Empirical results show that our optimizers can achieve favorable and comparable results across various architectures, including ResNets, Vision Transformers, and language models.
\end{itemize}

\textbf{Organization}. The rest of the paper is organized as follows: The next section will discuss related works on memory-efficient optimization. We provide in section~\ref{sec:background} a brief overview of gradient-based optimization and the Hamiltonian descent framework. Section~\ref{sec:hfac} presents our main methodological contributions. The experimental results are shown in section~\ref{sec:experiments}. Finally, in section~\ref{sec:discussion}, we will discuss the potential of our proposal for future works. The proofs, experimental settings, and additional results are left in Supplementary Materials.

\textbf{Notations}: We denote model parameters by a matrix of size $m \times n$. Let $\vv 1_m$ and $\vv 1_n$ be vectors of ones with dimensions $m$ and $n$, respectively. For any matrices $\vv A, \vv B$ of size $m \times n$, we use $\sqrt{\vv A}$ for element-wise square root, $\vv A^2$ for element-wise square, and $\vv A / \vv B$ to denote element-wise division. $\vv A^\top$ stands for the transpose matrix, $\text{trace}(\vv A)$ denotes the trace of \textit{square} matrix $\vv A$, and the function $\mathrm{RMS}(\vv A)=\sqrt{\frac{1}{m \times n} \sum_{i, j} A^2_{ij}}$ represents root-mean-square calculation.

\section{RELATED WORKS}

Numerous techniques have been developed to address memory overhead in optimization algorithms, each targeting specific aspects to enhance efficiency.

Adafactor~\citep{shazeer2018adafactor} minimizes the memory cost to a sublinear level by factorizing the second-moment estimator using a row-column outer product. Luo et al.~\citep{luo2023came} introduced a confidence-guided strategy to mitigate erroneous updates and reduce instability in Adafactor training, while still retaining the same memory footprint. SM3~\citep{anil2019memory} is another efficient algorithm, which organizes the parameter space into sets and simplifies the maintenance of second-moment statistics by computing the maximum squared gradient for each set. 

There are several methods that focus on the low-rank structure of the gradient rather than the moment statistics~\citep{gooneratne2020low, zhao2024galore}. For instance, GaLore~\citep{zhao2024galore} periodically applies Singular Value Decomposition (SVD) to project the full gradients onto a lower-dimensional subspace, and subsequently utilize these projected gradients in adaptive optimization processes. Sketchy~\citep{feinberg2024sketchy} leverages the Frequent Directions (FD) sketch technique to maintain a low-rank approximation of the gradient covariances matrix. Additionally, there is a line of work adapting quantization to reduce the memory cost of optimizer states~\citep{dettmers20218, dettmers2024qlora, li2024memory}.

For second-order (Hessian-based) methods, the memory constraints primarily arise from computing the inverse Hessian matrix. Some successful works have also resolved this bottleneck using Gauss-Newton decomposition~\citep{liu2023sophia} or Kronecker-based factorization~\citep{mozaffari2024mkor}.

\section{BACKGROUND} \label{sec:background}
\subsection{Gradient-based Optimization}
We consider an unconstrained, continuous optimization problem $\min_{\vv W \in \mathbb{R}^{m \times n}} f(\vv W)$,
with a proper differentiable and lower bounded objective $f: \mathbb{R}^{m \times n} \rightarrow \mathbb{R}$. In deep learning, gradient-based optimization algorithms have been the de facto choice for solving such problems, with first-order methods being widely adopted due to their scalability and effectiveness. Formally, we can unify these algorithms in a generalized form as:
\begin{equation}
    \vv W_{t+1} = \vv W_{t} + \varphi_t (\vv S_t) \quad \vv S_t = \gamma_t (\vv S_{t-1}, \nabla f(\vv W_t)) \label{eq:general-form}
\end{equation}
where $\vv S_t$ is the optimization state, and $\varphi_t, \gamma_t$ are some mapping functions. Principled designs of $\vv S_t$ rely on first-order gradient statistics, leading to popular algorithms including Momentum Gradient Descent (GDm~\citep{polyak1964some}), Sign Momentum (Signum~\cite{bernstein2018signsgd}, Lion~\citep{chen2024symbolic}), and adaptive optimizers like RMSprop~\citep{tieleman2012lecture}, Adam~\citep{kingma2014adam} and variants.

Specifically with Adam, the optimization states consist of estimates for the mean and uncentered variance of gradient information. These moment estimators are computed by applying exponential moving averages (EMA) on both gradients and their squares across training iterations. The update rules are:
\begin{align}
    & \quad \vv M_t = \hat{\beta}_{1t} \vv M_{t-1} + (1 - \hat{\beta}_{1t})\nabla f(\vv W_t)  \nonumber \\
    & \quad \vv V_t = \hat{\beta}_{2t} \vv V_{t-1} + (1 - \hat{\beta}_{2t})\nabla f(\vv W_t)^2  \\
    & \quad \vv W_{t+1} = \vv W_t - \eta_t \vv M_t / (\sqrt{\vv V_t} + \epsilon) \nonumber
\end{align}
where the decay moment coefficients $\hat{\beta}_{1,2t} = \beta_{1,2} (1-\beta^{t-1}_{1,2})/(1-\beta^{t}_{1,2})$ is defined as equivalent to the bias correction step. By leveraging the second moment $\vv V_t$, Adam can dynamically adjust the learning rate for each parameter, making it highly effective at navigating non-convex loss landscapes. However, this advantage comes with the trade-off of increased memory overhead.


\paragraph{Adafactor optimizer.} ~\cite{shazeer2018adafactor} address Adam's memory cost by employing an efficient rank-1 parameterization for the scaling factor, $\vv V \approx \vv r \vv s\tt $. The updates of vectors $r$ and $s$ were derived by minimizing the total elementwise I-divergence subject to componentwise non-negative constraints: 
$$
\underset{\vv r \in \mathbb{R}^m, \vv s \in \mathbb{R}^n}{\text{minimize}} \sum_{i=1}^m \sum_{j=1}^n d(V_{ij}, r_is_j),
$$
in which $r_i \geq 0, s_j \geq 0$ and $d(p,q) = p \log \frac{p}{q} - p + q$.
\begin{algorithm}[t]
\caption{\textcolor{magenta}{Adafactor-m} for matrix parameters.}
\label{alg:algorithm1}
\begin{algorithmic}
\STATE \textbf{Inputs:} moment decay coefficients $\beta_1$, $\beta_2$, smoothing term $\epsilon$, and regularization constant $\lambda$
\STATE \textbf{Initialization:} weight parameters $\vv W_1 \in \mathbb{R}^{m \times n}$, initial moments $\vv M_0, \vv r_0, \vv s_0 \leftarrow 0$
\FOR{$t=1$ to $T$}
    \STATE $\vv G_t = \nabla f_t(\vv W_{t})$
    \STATE $\vv M_t = \hat{\beta}_{1t} \vv M_{t-1} + (1-\hat{\beta}_{1t}) \vv G_t$
    \STATE $\vv r_t = \hat{\beta}_{2t} \vv r_{t-1} + (1-\hat{\beta}_{2t}) \big[(\vv G_t)^2 + \epsilon \big] \vv 1_n$
    \STATE $\vv s_t = \hat{\beta}_{2t} \vv s_{t-1} + (1-\hat{\beta}_{2t}) \big[(\vv G_t^\top)^2 + \epsilon \big] \vv 1_m$
    \STATE $\widehat{\vv V}_t = \vv r_t \vv s_t^\top / (1_m^\top \vv r_t)$
    \STATE $\vv W_{t+1} = \vv W_{t} - \eta_t \left( \clip \left(\vv M_t / \sqrt{\widehat{\vv V}_t} \right) + \lambda \vv W_{t} \right)$
\ENDFOR
\end{algorithmic}
\end{algorithm}
Solving this problem results in a closed-form solution denoted by $\vv r = \vv V \vv 1_m, \vv s = \vv V^\top \vv 1_n / \vv r^\top \vv 1_m$. Subsequently, the iterative process of Adafactor optimizer can be outlined as in Algorithm ~\ref{alg:algorithm1}. In the model parameter update, we have $\clip(\vv U) = \vv U / \max (1, \mathrm{RMS}(\vv U)/d)$ with $\mathrm{RMS}(.)$ refers to the root-mean-square function and $d$ is the threshold value, meaning that we cap the norm of the actual update rather than just the gradient. This technique aims to eliminate the larger-than-desired updates and also stabilize the training process with slow decay. Briefly, Adafactor tracks the moving averages of the row and column sums of squared gradients over iterations, resulting in estimates of factored second moment $\vv r_t$ and $\vv s_t$. A normalized outer product $\vv r_{t} \vv s_{t}^\top / (\vv 1_m^\top \vv r_{t})$ is then used to reconstruct a low-rank parameterization of the second moment $\widehat{\vv V}_t$.

\subsection{Hamiltonian Descent}
One powerful approach to studying the dynamic behavior of the gradient-based optimization methods is to examine their continuous-time forms in the limit of infinitesimal step size~\citep{maddison2018hamiltonian, gao2022global, chen2023lion}. In this regime, we can derive insights into the asymptotic convergence of the algorithms, abstracting away the choices of step size, discretization, and accumulation errors. 

Let's examine the case of Momentum Gradient Descent (GDm) for simplicity. The update procedure of GDm is based on the following iterative scheme:
\begin{align}
    \vv W_{t+1} = \vv W_{t} - \eta_t \vv M_t, \quad \vv M_t = \beta \vv M_{t-1} + \nabla f(\vv W_{t})
\end{align}
In case we employ \textit{full-batch} gradients, the GDm scheme becomes deterministic and can be viewed as a discretization of a \emph{continuous-time system} as follows:
\begin{align}\label{eq:fullm-ode}
    \dfrac{\mathrm{d}}{\mathrm{d}t} \vv W_t = \vv M_t, \quad \quad  \dfrac{\mathrm{d}}{\mathrm{d}t} \vv M_t = -\gamma {\vv M_t} - \nabla f(\vv W_t)
\end{align}
where $\vv M_t$ and $\gamma$ are in analogy to the \emph{velocity} and \emph{friction} in classical mechanics. In principle, this ordinary differential equation (ODE) defines a trajectory of the particle $\vv W_t$ and its velocity $\vv M_t$, which drives the system characterized by the \textit{total energy} or Hamiltonian function,
$$
\mathcal{H}(\vv W, \vv M) = f(\vv W) + \|\vv M\|_2^2/2,
$$
towards stationary points.  The Hamiltonian $\mathcal{H}(.)$ is an augmented function of the original objective $f(.)$ and it satisfies
$
    \min_{\vv M} \mathcal{H}(\vv W, \vv M) = \mathcal{L}(\vv W) \, \forall \vv W,
$
meaning that minimizing $\mathcal{H}(\vv W, \vv M)$ will reduce to minimizing $f(\vv W)$.
\noindent
\paragraph{Convergence to Local Optima.}We can show that the Hamiltonian $\mathcal{H}_t$ is monotonically decreasing along the ODE trajectory:
\begin{align*}
    \dfrac{\mathrm{d}}{\mathrm{d}t} \mathcal{H}_t &= \text{trace}(\nabla^\top f(\vv W_t) \dfrac{\mathrm{d}}{\mathrm{d}t} \vv W_t) + \text{trace}(\vv M_t^\top \dfrac{\mathrm{d}}{\mathrm{d}t} \vv M_t) \\
    &= -\gamma \text{trace}(\vv M_t^\top \vv M_t) = -\gamma \|\vv M_t \|^2_F \leq 0.
\end{align*}
with $\norm{.}_F$ is the Frobenius norm and $\mathcal{H}_t \triangleq \mathcal{H}(\vv W_t, \vv M_t)$. By LaSalle’s Invariance principle, the set of accumulation points $(\vv W_t, \vv M_t)$ must be contained in $\mathcal{I}$, where $\mathcal{I} = \{\text{the union of complete trajectories satisfying }$ $\dfrac{\mathrm{d}}{\mathrm{d}t} \mathcal{H}(\vv W_t, \vv M_t) = 0, \forall t \}$. The points in the limit set $\mathcal{I}$ should satisfy $\vv M_t \equiv 0$, leading to $\dfrac{\mathrm{d}}{\mathrm{d}t} \vv W_t \equiv 0$ and $\nabla f(\vv W_t) \equiv 0$. This indicates that all trajectories will converge to local optimal points.

The Hamiltonian interpretation is meaningful because its underlying logic suggests that a proper optimizer should be guaranteed to converge to the local optima of loss function, at least when the step sizes are sufficiently small. This powerful tool can facilitate us to theoretically establish a broader class of optimizers with guarantees, as well as providing complementary insights, improving or extending existing algorithms. In the next section, we will reinforce this perspective by extending the Hamiltonian framework to more modern algorithms, especially by providing new memory-efficient optimizers based on this principle.

\section{FACTORIZED HAMILTONIAN DESCENT}
\label{sec:hfac}
In this section, we first interpret Adafactor optimizer from the perspective of Hamiltonian descent. We then introduce a general approach to factorize the momentum in first-order optimization methods. We will specifically deliver a factorization version for the sign-based momentum update and name this optimizer as \textit{sign}FSGD. Based on further insights, we propose a novel adaptive factorized optimization algorithm, named \ours{}. This is our main algorithmic contribution to this paper.

\subsection{Adafactor optimizer as Hamiltonian}
The factorization technique used in Adafactor is computationally efficient and scalable as it offers analytical formulations without the need for additional approximations. However, directly applying a non-negative decomposition approach like that often appears heuristic. It fails to provide any insights into the optimization dynamics and convergence guarantees when we reparameterize the second-moment estimator using such a low-rank representation. Compared to simple optimizers like momentum gradient descent, Adafactor involves complex interactions between model parameter $\vv W_t$ and optimization states $\{\vv M_t, \vv r_t, \vv s_t\}$. This could hinder the analysis of convergence or other theoretical properties.

Interestingly, we demonstrate that the iterative procedure in Adafactor-m algorithm can be discretized from an ordinary differential equation (ODE) as follows:
\begin{align}
    \dfrac{\mathrm{d}}{\mathrm{d}t} \vv W_t &= -\frac{\vv M_t}{\sqrt{\vv r_t \vv s_t^\top / \vv 1_m^\top \vv r_t}}, & \dfrac{\mathrm{d}}{\mathrm{d}t} \vv M_t &= \vv G_t - \alpha \vv M_t \\
    \dfrac{\mathrm{d}}{\mathrm{d}t} \vv r_t &= (\vv G_t)^2 \vv 1_n - \alpha \vv r_t, & \dfrac{\mathrm{d}}{\mathrm{d}t} \vv s_t &= (\vv G_t^\top)^2 \vv 1_m - \alpha \vv s_t \nonumber
    \tag*{\textcolor{magenta}{// Adafactor-m (ODE)}}
\end{align}
which solves a minimization problem with respect to the Hamiltonian function described by:
$$
\mathcal{H}(\vv W, \vv M, \vv r, \vv s) = f(\vv W) + \frac{1}{2} \sum_{i=1, j=1}^{m, n} \dfrac{ M_{ij}^2 \sqrt{\sum_{i=1}^m r_i}}{\sqrt{ r_i s_j}}.
$$ 
\begin{proposition}\label{pro:adafactor}
A key property is that the function $\mathcal{H}$ monotonically decreases along the ODE trajectory, that is, $\dfrac{\mathrm{d}}{\mathrm{d}t} \mathcal{H}(\vv W_t, \vv M_t, \vv r_t,\vv s_t) \leq 0$.
\end{proposition}
\noindent
A detailed proof is provided in Appendix~\ref{app:adafactor}. 

Proposition~\ref{pro:adafactor} is a natural extension of the traditional Hamiltonian Descent presented in Section~\ref{sec:background}, where the regularization term in the Hamiltonian function $\mathcal{H}(\vv W, \vv M, \vv r, \vv s)$ still plays a role as kinetic energy. The key difference is that the momentum $\vv M$ is now normalized by the scaling parameter $\vv r \vv s^\top / (1_m^\top \vv r)$, making the term equivalent to the kinetic energy with non-uniform mass values.
In the next sections, we will broaden the framework to encompass a wider range of optimizers, including both momentum-based and adaptive algorithms. By doing so, we aim to integrate various memory-efficient optimization techniques within a unified framework of Hamiltonian dynamics.

\begin{figure*}[t]
    \centering
    \begin{subfigure}[b]{0.51\textwidth}
        \centering
        \includegraphics[width=0.8\textwidth]{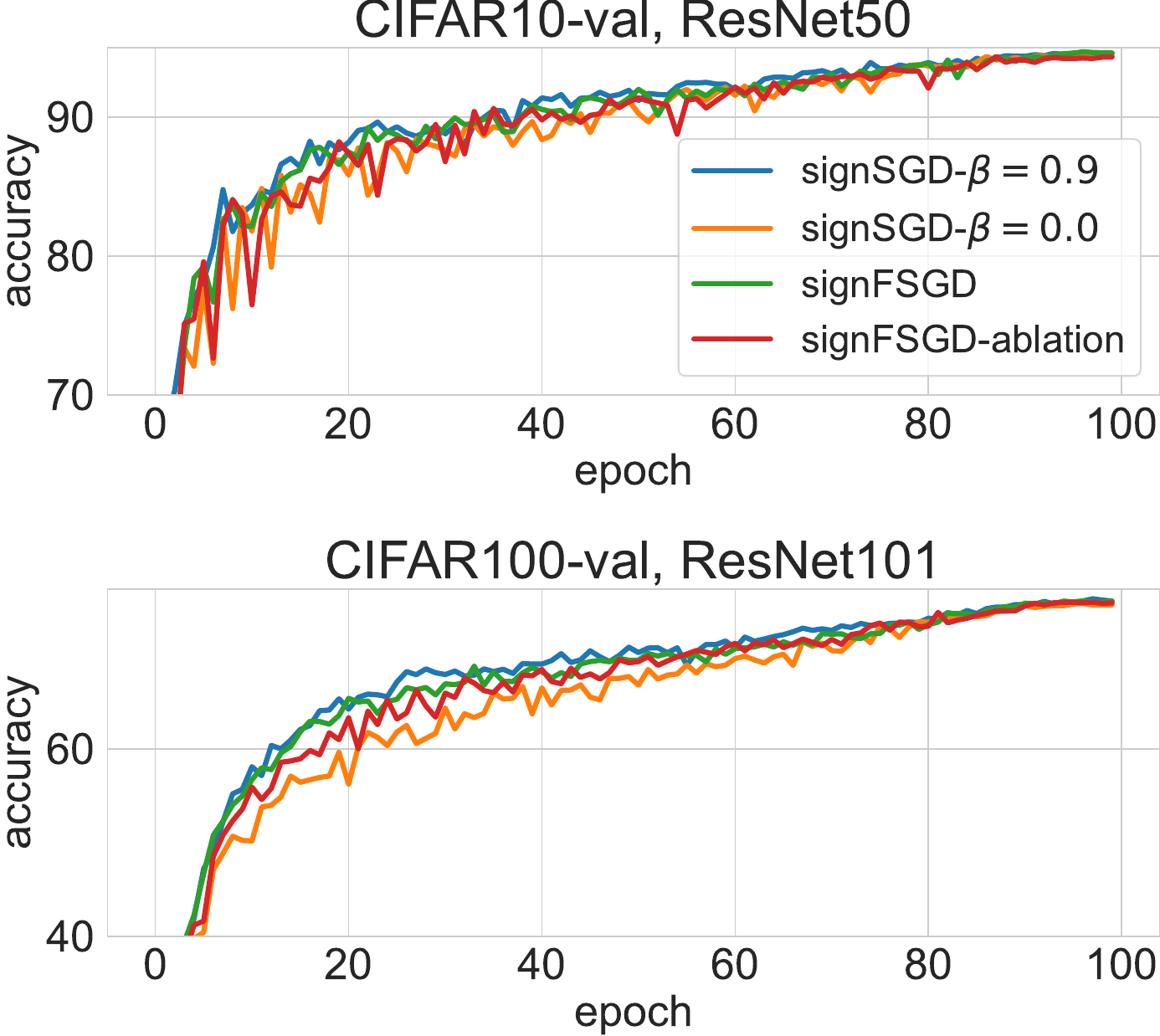}
        \caption{Validation accuracy on CIFAR datasets}
        \label{fig:cifar-resnet}
    \end{subfigure}
    \hfill
    \begin{subfigure}[b]{0.46\textwidth}
        \centering
        \includegraphics[width=0.8\textwidth]{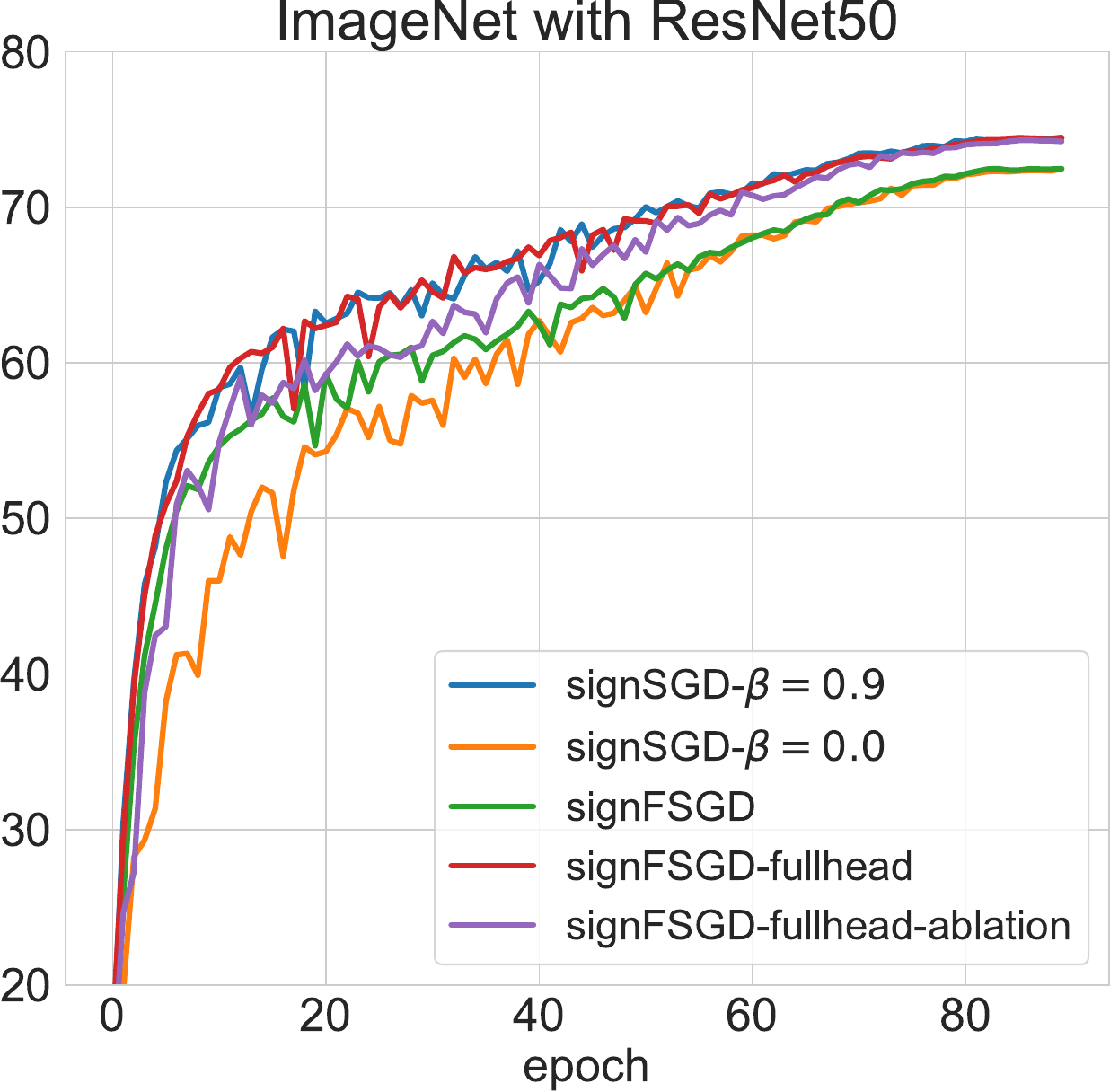}
        \caption{Validation accuracy on ImageNet1K}
        \label{fig:imagenet-resnet}
    \end{subfigure}
    \caption{A comparison of optimizer performance on ResNet architectures. For \textit{sign}SGD, $\beta$ denotes the momentum coefficient. For \textit{sign}FSDG, ``\textit{ablation}'' means the version without corrected terms, ``\textit{fullhead}'' means the version using full momentum for the MLP head layer.}
    \label{fig:resnet}
\end{figure*}

\subsection{Factorized first-moment estimation}\label{sec:first-moment-fac}
Factoring the second moment is a straightforward yet intuitive idea to handle memory overhead. This approach offers an advantage in that it allows for the tolerance of some information loss, as our focus is solely on the magnitude of the scaling parameter. Despite this, most algorithms still exhibit an inherent drawback of requiring a full momentum to avoid performance degradation. Dealing with the first moment of gradient information, however, poses more challenging problems, as we must consider both the magnitude and update directions.

\textbf{Factorization via rank-1 parameterization}. We propose here a sublinear memory optimizer that factorizes the first-moment estimator. The algorithm is specifically derived from an ordinary differential equation outlined as (recall that $\vv W \in \RR^{m\times n}$ is a matrix): 
\begin{align} \label{eq:fac1-ode}
    \dfrac{\mathrm{d}}{\mathrm{d}t} \vv W_t &= 
    -\nabla \phi \big( \beta \hat{\vv u}_t \vv 1_n^\top + \vv G_t \big)  
    -\nabla \psi \big( \beta \vv 1_m \hat{\vv v}_t^\top + \vv G_t \big) \nonumber \\ 
    \dfrac{\mathrm{d}}{\mathrm{d}t} \vv u_t &= \vv G_t \vv 1_n / n - \alpha \vv u_t, \quad \hat{\vv u}_t = \vv u_t - \vv G_t \vv 1_n / n \\    
    \dfrac{\mathrm{d}}{\mathrm{d}t} \vv v_t &= \vv G_t^\top \vv 1_m / m - \alpha \vv v_t, \quad \hat{\vv v}_t = \vv v_t - \vv G_t^\top \vv 1_m / m \nonumber
    \tag*{\textcolor{magenta}{// Factorized first moment (ODE)}}
\end{align}
where $\alpha,\beta$ are two positive parameters; $\phi$ and $\psi$ are any convex functions, such that $\dd\phi(\cdot), \dd\psi(\cdot)$ are monotonic operators.
The simplest option to consider is that $\phi(\vv W) = \psi(\vv W) = \norm{\vv W}_2^2/2$, then their gradients $\dd\phi(\vv W)=\dd\psi(\vv W) = \vv W$. This system yields the Hamiltonian function: 
$$
\mathcal{H}(\vv W, \vv u, \vv v) = f(\vv W) + \frac{\beta n}{2} \norm{\vv u}_2^2 + \frac{\beta m}{2} \norm{\vv v}_2^2.
$$
A detailed analysis is provided in Appendix~\ref{app:first-order}.

\begin{algorithm}[t]
\caption{\textcolor{magenta}{\textit{sign}FSGD} for matrix parameters.}
\label{alg:signfsgd}
\begin{algorithmic}
\STATE \textbf{Inputs:} moment coefficients $\beta=0.9$, and regularization constant $\lambda$
\STATE \textbf{Initialization:} weight parameters $\vv W_1 \in \mathbb{R}^{m \times n}$, initial moment factors $\vv u_0, \vv v_0 \leftarrow 0$
\FOR{$t=1$ to $T$}
    \STATE $\vv G_t = \nabla f_t(\vv W_{t})$
    \STATE $\vv u_{t} = \beta \vv u_{t-1} + (1-\beta) \vv G_t \vv 1_n / n $
    \STATE $\vv v_{t} = \beta \vv v_{t-1} + (1-\beta) \vv G_t^\top \vv 1_m / m $
    \STATE $\hat{\vv u}_{t} = \vv u_t - \vv G_t \vv 1_n / n $
    \STATE $\hat{\vv v}_{t} = \vv v_t - \vv G_t^\top \vv 1_m / m $
    \STATE $\vv W_{t+1} = \vv W_{t} - \eta_{t} \Big( \sign (\beta \hat{\vv u}_{t} \vv 1_n^\top + \vv G_t) $ 
    \STATE \hspace{2.5cm} $+\ \sign (\beta \vv 1_m \hat{\vv v}_{t}^\top + \vv G_t) + \lambda \vv W_{t} \Big)$
\ENDFOR
\end{algorithmic}
\end{algorithm}

\begin{figure*}[ht]
    \centering
    \includegraphics[width=0.85\textwidth]{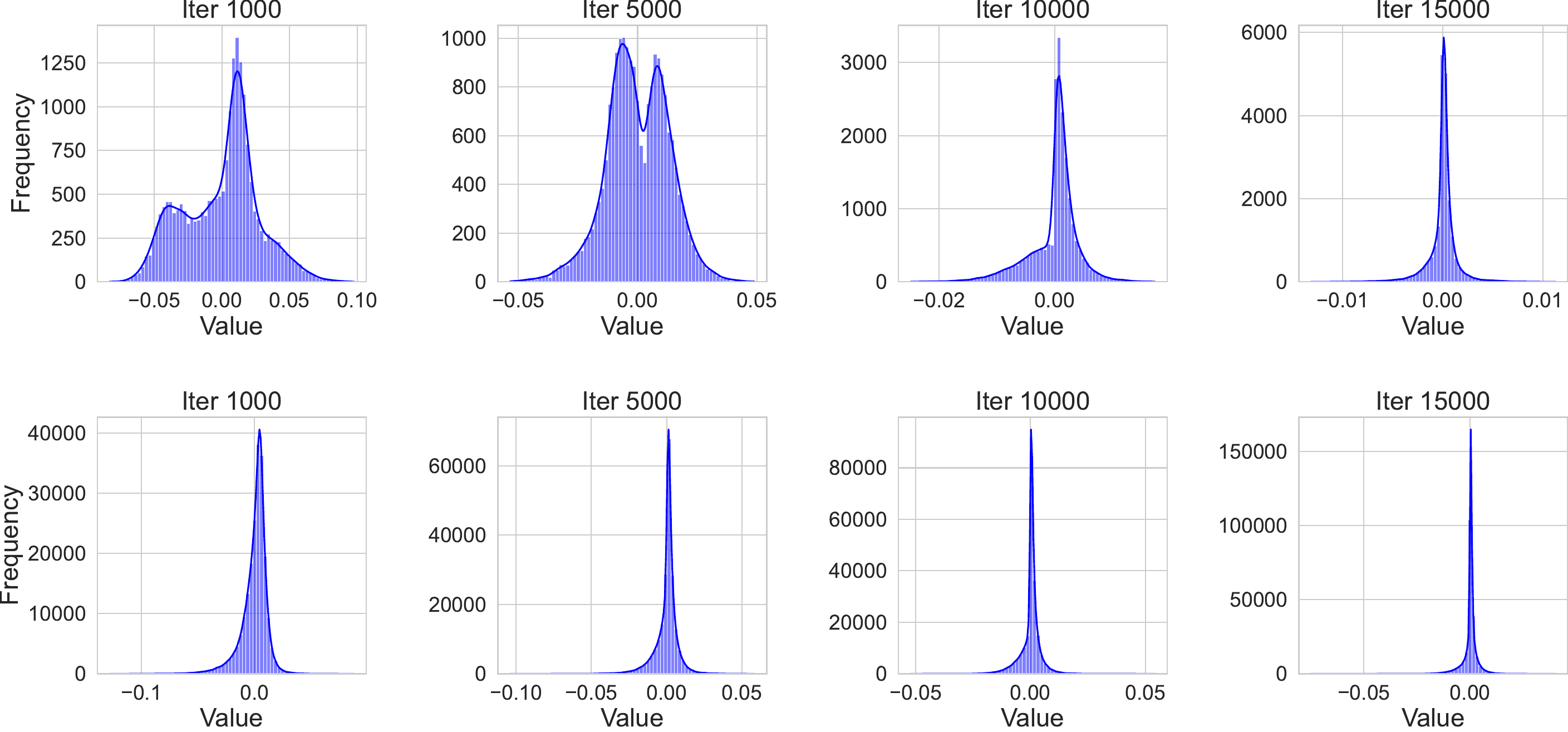}
    \caption{Histograms illustrating the gradients of MLP head layers in ResNet50 (top) and ResNet101 (bottom) trained on CIFAR10 and CIFAR100, respectively.}
    \label{fig:grad_hist}
\end{figure*}

\textbf{Back to our framework, a number of remarks are in order}: 

\noindent
1) Compared with algorithms that maintain a full rank momentum matrix $\vv M \in \RR^{m\times n}$ as defined in (\ref{eq:fullm-ode}), 
this algorithm employs two rank-one momentum vectors $\vv u\in \RR^{m}$ and $\vv v \in \RR^{n}$, 
which significantly reduces the memory cost from $O(mn) $ to $O(m+n)$. 
In the algorithm, we use the simple unit vectors $\vv 1_m, \vv 1_n$ to first down project the full gradient $\dd f(\vv W)$ 
into the rank-one spaces of column means and row means, and exponentially accumulate these statistics across training steps.  We then reconstruct the rank-one momentum $\hat {\vv u}, \hat {\vv v}$ to update the model parameters. The $\hat {\vv u}$ and $\hat {\vv v}$ are centralized variants of $\vv u$ and $\vv v$.


\noindent
2) Discretizing ODE (\ref{eq:fac1-ode}) using the Euler method results in the following iterative scheme: 
\begin{align*}
    \vv u_{t} &= \hat{\beta}_{1t} \vv u_{t-1} + (1-\hat{\beta}_{1t}) \vv G_t \vv 1_n /n , ~~\hat{\vv u}_t = \vv u_t - \vv G_t \vv 1_n/n\\
    \vv v_{t} &= \hat{\beta}_{1t} \vv v_{t-1} + (1-\hat{\beta}_{1t}) \vv G_t^\top \vv 1_m /m, ~~\hat{\vv v}_t = \vv v_t - \vv G_t^\top \vv 1_m/m \\
    \vv W_{t} &= \vv W_{t-1} -\eta_{t}\bigg[ \nabla \phi \left(\hat{\beta}_{1t} \hat{\vv u}_{t} \vv 1_n^\top   + \vv G_t \right) \\
    & \hspace{3cm} + \nabla \psi \left(\hat{\beta}_{1t} \vv 1_m \hat{\vv v}_{t}^\top  + \vv G_t  \right) \bigg]  
\end{align*}
In discrete-time analysis, we centralize the column-means and row-means statistics $\vv u_t, \vv v_t$ by corrected terms $\vv G_t \vv 1_n/n$ and $\vv G_t^\top \vv 1_m/m $, respectively. These corrected terms are essential as they can guarantee the updates inside $\nabla \phi(.)$ and $\nabla \psi(.)$ will converge to the true first moment $\mathbb{E}[\vv G_t]$. Specifically, by unfolding the factored moment $\vv u_t$ (and $\vv v_t$, similarly), we can get an accumulator of column-wise mean of gradients as:
$$
\vv u_t = \sum_{\tau=1}^t(1-\hat{\beta}_{1\tau})\prod_{\kappa=\tau+1}^t \hat{\beta}_{1\kappa} \vv G_\tau \vv 1_n.
$$
By the assumption that the gradient distribution is stationary, taking expectations gives us: 
\begin{align*}
    \mathbb{E}[\vv u_t] &= \sum_{\tau=1}^t(1-\hat{\beta}_{1\tau})\prod_{\kappa=\tau+1}^t \hat{\beta}_{1\kappa} \mathbb{E}[\vv G_\tau \vv 1_n] \\
    &= \sum_{\tau=1}^t(1-\hat{\beta}_{1\tau})\prod_{\kappa=\tau+1}^t \hat{\beta}_{1\kappa} \mathbb{E}[\vv G_t \vv 1_n].
\end{align*}
Futhermore, we can prove by induction that $\sum_{\tau=1}^t(1-\hat{\beta}_{1\tau})\prod_{\kappa=\tau+1}^t \hat{\beta}_{1\kappa} = 1$, leading to $\mathbb{E}[\vv u_t] = \mathbb{E}[\vv G_t) \vv 1_n]$ and hence the expected value of centralized moment factor $\mathbb{E}[\hat{\vv u}_t]$ will approximate the true first moment $\mathbb{E}[\vv G_t]$.

\noindent
3) When the momentum coefficient is deactivated ($\beta_1=0$), the update terms inside $\nabla \phi(.)$ and $\nabla \psi(.)$ degenerate to $\vv G_t$. In this scenario, the algorithm simplifies to standard gradient descent:
$$\dfrac{\mathrm{d}}{\mathrm{d}t} \vv W_t = - \dd\phi(\vv G_t) - \dd \psi(\vv G_t).$$ 
To empirically examine the prospects of our general framework and especially the necessity of the corrected terms, we conducted a basic image classification experiment on the CIFAR10 and CIFAR100 datasets using ResNet50 and ResNet101, respectively. We choose $\phi(.)$ and $\psi(.)$ to be the $L_1$-norm, namely $\phi(\vv W) = \psi(\vv W) = \norm{\vv W}_1$, resulting in their gradients $\nabla \phi(.)$ and $\nabla \psi(.)$ being $\textit{sign}(.)$ functions. In this case, we refer to our algorithm as \textcolor{magenta}{\textit{sign}FSGD}~\ref{alg:signfsgd}, which incorporates a rank-1 parameterization to factorize the momentum in \textit{sign}SGD optimizer~\citep{bernstein2018signsgd}. We will evaluate the effectiveness of \textit{sign}FSGD with and without the corrected terms, compare its performance to that of \textit{sign}SGD with momentum enabled or disabled. Figure~\ref{fig:cifar-resnet} shows that \textit{sign}FSGD can yield comparable results to \textit{sign}SGD with momentum. However, without the corrected terms, our algorithm is generally less stable and only gains relative improvements compared to \textit{sign}SGD without momentum. Thus, simply accumulating row means and column means of gradient information across training steps is insufficient to accelerate the optimization process. 

Motivated by the recent development of Lion optimizer~\citep{chen2024symbolic}, we adopted an efficient technique called double$-\beta$ scheme to further improve the performance of \textit{sign}FSGD on ResNet models. We refer to this new algorithm as \textcolor{magenta}{LionFactor}~\ref{alg:lionfactor}, and provide some discussions in Appendix~\ref{app:lionfactor}.

\subsection{Fully factorized moment estimations} \label{sec:full-factor}
\textbf{Inadequacy of first-moment factorization}. For ResNet models, mini-batch gradients are typically quite small and well concentrated around zero mean as shown in Figure~\ref{fig:grad_hist}. Consequently, accumulating the momentum through column means and row means of the gradient information will significantly flatten the magnitude. This issue could potentially reduce the impact of momentum on the optimization updates, especially with large-scale settings such as ImageNet and modern deep learning architectures. 

In Figure~\ref{fig:imagenet-resnet}, we present the performances of \textit{sign}SGD and \textit{sign}FSGD when training ResNet50  from scratch on ImageNet1K. Notably, our method does not yield substantial enhancements compared to \textit{sign}SGD without momentum as observed on CIFAR10, CIFAR100 datasets. However, when employing a full momentum instead of the factorized one for the MLP head layer, our \textit{sign}FSGD can perform comparably to \textit{sign}SGD with momentum. We witness the same behavior when applying the algorithm to larger ResNet architectures.

This observation reinforces our previous remark regarding the moderate impact of our factorized first moment on scale models. Although a few minor customizations, such as omitting the factorization of the MLP head layer like above, could overcome the issue on ResNets, our initial expectation leaned towards a more general-purpose optimization algorithm. We specifically aim to extend the scalability of our algorithms to encompass modern convolution-free architectures.

\begin{figure*}[ht]
    \centering
    \includegraphics[width=0.93\textwidth]{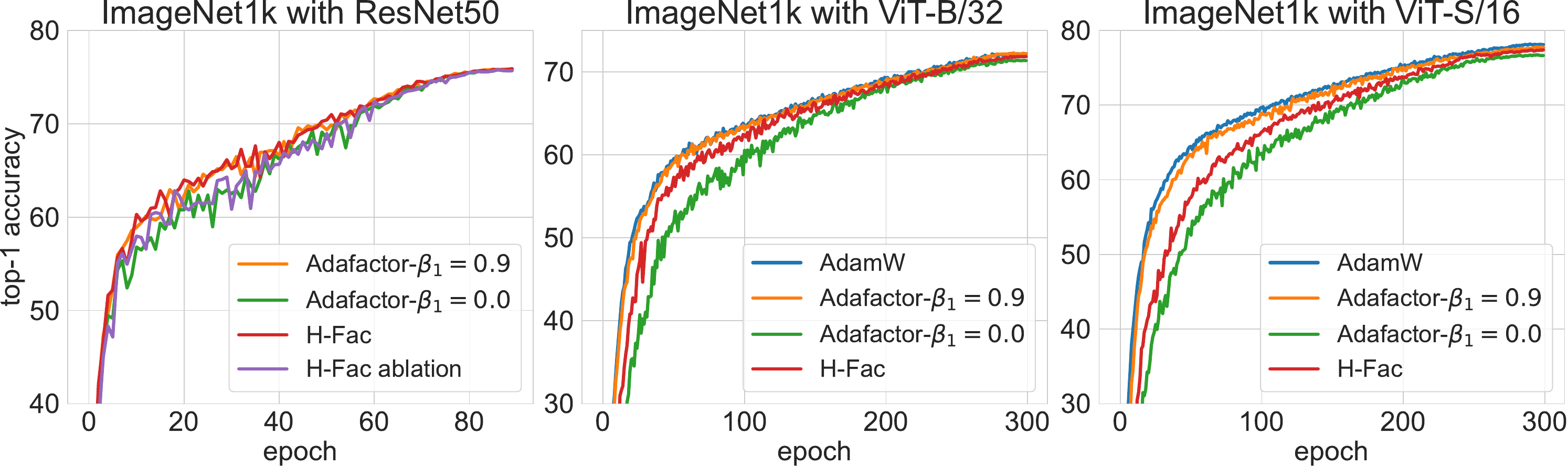}
    \caption{Top-1 Accuracy of optimizers in training ResNet50, ViT-B/32, and ViT-S/16 from scratch on the ImageNet1K. For \ours{}, ``ablation'' means the version without corrected terms.}
    \label{fig:imagenet_vit}
\end{figure*}

\textbf{Incorporating second-moment estimator}. A natural idea to deal with the magnitude issue of factorized first-moment estimator is to normalize them by factored second moments. Indeed, by integrating our factorization approach for the first-moment estimator into Adafactor optimizer, we can establish a unified method where both moment estimators are fully factorized. In principle, we aim to conceptualize our algorithms within Hamiltonian frameworks, but deriving an objective function for such a simplistic integration might be challenging. We instead come up with a novel algorithm characterized by the following ODE:
\begin{align*}
    \dfrac{\mathrm{d}}{\mathrm{d}t} \vv W_t &= - \frac{\vv G_t}{\sqrt{\vv r_t \vv s_t^\top / \vv 1_m^\top \vv r_t}} - \frac{1}{2} \left[\frac{\vv u_t \vv 1_n^\top - \vv G_t \vv 1_n \vv 1_n^\top / n }{\sqrt{\vv r_t \vv 1_n^\top }} \right.  \nonumber \\
    & \hspace{3.5cm} \left. + \frac{\vv 1_m \vv v_t^\top - \vv 1_m \vv 1_m^\top \vv G_t / m }{\sqrt{\vv 1_m \vv s_t^\top}} \right]  \nonumber \\ 
    \dfrac{\mathrm{d}}{\mathrm{d}t} \vv u_t &= \vv G_t \vv 1_n / n - \alpha \vv u_t, \quad \quad \dfrac{\mathrm{d}}{\mathrm{d}t} \vv v_t = \vv G_t^\top \vv 1_m / m - \alpha \vv v_t \\    
    \dfrac{\mathrm{d}}{\mathrm{d}t} \vv r_t &= (\vv G_t)^2 \vv 1_n  - \alpha \vv r_t, \quad \quad \dfrac{\mathrm{d}}{\mathrm{d}t} \vv s_t = (\vv G_t^\top)^2 \vv 1_m - \alpha \vv s_t \nonumber \nonumber   
    \tag*{\textcolor{magenta}{//  \ours{} (ODE)}}\label{equ::fac-2}
\end{align*}
which yields the following Hamiltonian function:
$$
\mathcal{H}(\vv W, \vv u, \vv v, \vv r, \vv s) := f(\vv W) + \frac{n}{4} \sum_{i=1}^m \frac{u_i^2}{\sqrt{r_i}} + \frac{m}{4} \sum_{j=1}^n \frac{v_j^2}{\sqrt{s_j}}.
$$
The discrete-time equivalent is presented in Algorithm~\ref{alg:algorithm2}. 
Our model parameters update can be seen as an accumulator of normalized gradients, where both the momentum and the current gradient are rescaled by their corresponding cumulative second-moment information. This formulation is distinct from Adam, which uses the signal-to-noise ratio $\vv M_t / \sqrt{\vv V_t}$ to update the parameters. More specifically, our update includes a combination of three key elements: (i) the normalized momentum factorization $0.5*(\phi_{term} + \psi_{term})$, in which the factorized first moment in section~\ref{sec:first-moment-fac} is normalized by the row means and the column means of second-moment estimators; (ii) a clipping of normalized gradient $ \clip \big (\vv G_t / \sqrt{\hat{\vv V}_{t}} \big)$ which is inherited from Adafactor update; and (iii) a decouple weight decay $ \lambda \vv W_{t}$ for enhancing the generalization performance of adaptive optimizers~\citep{loshchilov2017decoupled}.

\begin{algorithm}[t]
\caption{\textcolor{magenta}{\ours{}} for matrix parameters.}
\label{alg:algorithm2}
\begin{algorithmic}
\STATE \textbf{Inputs:} moment decay coefficients $\beta_1, \beta_2$, smoothing term $\epsilon$, and regularization constant $\lambda$
\STATE \textbf{Initialization:} weight parameters $\vv W_1 \in \mathbb{R}^{m \times n}$, initial factored moments $\vv u_0, \vv v_0, \vv r_0, \vv s_0 \leftarrow 0$
\FOR{$t=1$ to $T$}
    \STATE $\vv G_t = \nabla f_t(\vv W_{t})$
    \STATE $\vv u_{t} = \hat{\beta}_{1t} \vv u_{t-1} + (1-\hat{\beta}_{1t}) \vv G_t \vv 1_n / n$
    \STATE $\vv v_{t} = \hat{\beta}_{1t} \vv v_{t-1} + (1-\hat{\beta}_{1t}) \vv G_t^\top \vv 1_m / m$
    \STATE $\vv r_{t} = \hat{\beta}_{2t} \vv r_{t-1} + (1-\hat{\beta}_{2t}) \big[(\vv G_t)^2 + \epsilon \big] \vv 1_n$
    \STATE $\vv s_{t} = \hat{\beta}_{2t} \vv s_{t-1} + (1-\hat{\beta}_{2t}) \big[(\vv G_t^\top)^2 + \epsilon \big] \vv 1_m$
    \STATE $\widehat{\vv V}_{t} = \vv r_{t}\vv s_{t}^\top / (\vv 1_m^\top \vv r_{t})$
    \STATE $\phi_{term} = \hat{\beta}_{1t} \big(\vv u_{t} \vv 1_n^\top - \vv G_t \vv 1_n \vv 1_n^\top / n \big) / \sqrt{{\textcolor{blue}{\vv r_{t} \vv 1_n^\top /n}}}$
    \STATE $\psi_{term} = \hat{\beta}_{1t} \big(\vv 1_m \vv v_{t}^\top - \vv 1_m \vv 1_m^\top \vv G_t / m \big) / \sqrt{\textcolor{blue}{\vv 1_m \vv s_{t}^\top /m}}$
    \STATE $\vv W_{t+1} = \vv W_{t} - \eta_{t} \Big( 0.5(\phi_{term} + \psi_{term}) $
    \STATE \hspace{2.5cm} $ +\ \clip \left( \vv G_t / \sqrt{\textcolor{blue}{\widehat{\vv V}_{t}}} \right) + \lambda \vv W_{t} \Big)$%
\ENDFOR
\end{algorithmic}
\end{algorithm}

\section{EXPERIMENTS} \label{sec:experiments}

In this section, we conduct several experiments to demonstrate the effectiveness of our \ours{} algorithm. Unlike the earlier proposal of \textit{sign}FSGD, we do not customize the algorithm for any specific layers, but perform factorization on the entire architectures.

\subsection{Image Classification}
\textbf{Setup.} We evaluated the optimization algorithms described in this paper on pre-training ResNet50 and Vision Transformers (ViTs) from scratch using ImageNet1K dataset, following previous works~\citep{dosovitskiy2020image, he2016deep}. The images are pre-processed by Inception-style cropping~\citep{szegedy2016rethinking} and random horizontal. We train ResNet50 for 90 epochs, using a batch size of 1024, with cosine learning rate decay scheduler. For ViTs, we train them for 300 epochs, using a batch size of 4096, with a learning rate schedule of 10,000 steps warmup followed by linear decay. We also adopted strong data augmentations, including RandAugment(2,15)~\citep{Cubuk2019RandaugmentPA} and mixup (0.5)~\citep{Zhang2017mixupBE}, to boost ViTs performances. Hyperparameters tuning of learning rate, weight decay, and dropout rate is given in Appendix~\ref{app:hyper-settings}.
\paragraph{Evaluation.} The results are shown in Figure~\ref{fig:imagenet_vit} and Table~\ref{tab:imagenet}. On ResNet50, we can notice consistent patterns with those in Figure~\ref{fig:imagenet-resnet}. While our algorithm, \ours{}, can achieve similar memory efficiency as Adafactor without momentum, it demonstrates more stable training and delivers highly competitive performance compared to Adafactor with momentum. The performance degradation is also clearly evident in the ablation study, consolidating the significance of the corrected terms in our algorithms. On ViT models, despite performance gaps in the early stages, \ours{} progressively reaches the level of AdamW and Adafactor with momentum as training advances. Furthermore, the factorized momentum in our algorithm actually accelerates the optimization process, as evidenced by substantial improvements over Adafactor without momentum. 


\subsection{Language Modeling}
\textbf{Setup.} We apply the algorithms to pre-train LLaMA-based large language model models~\citep{zhang2019root, shazeer2020glu, touvron2023llama} on the C4 dataset~\citep{raffel2020exploring}. We measured the perplexity of the models on the validation set throughout training to assess convergence properties and final model performance. Specifically, we trained LLaMA models of sizes 60M and 130M for 10K and 20K, respectively. The learning rate schedule included a warmup phase during the first 10\% of the training steps, followed by cosine annealing that decayed the learning rate to 10\% of its initial value. All models use a maximum sequence length of 256 and a batch size of 512.

\begin{figure}[t]
    \centering
    \includegraphics[width=0.5\textwidth]{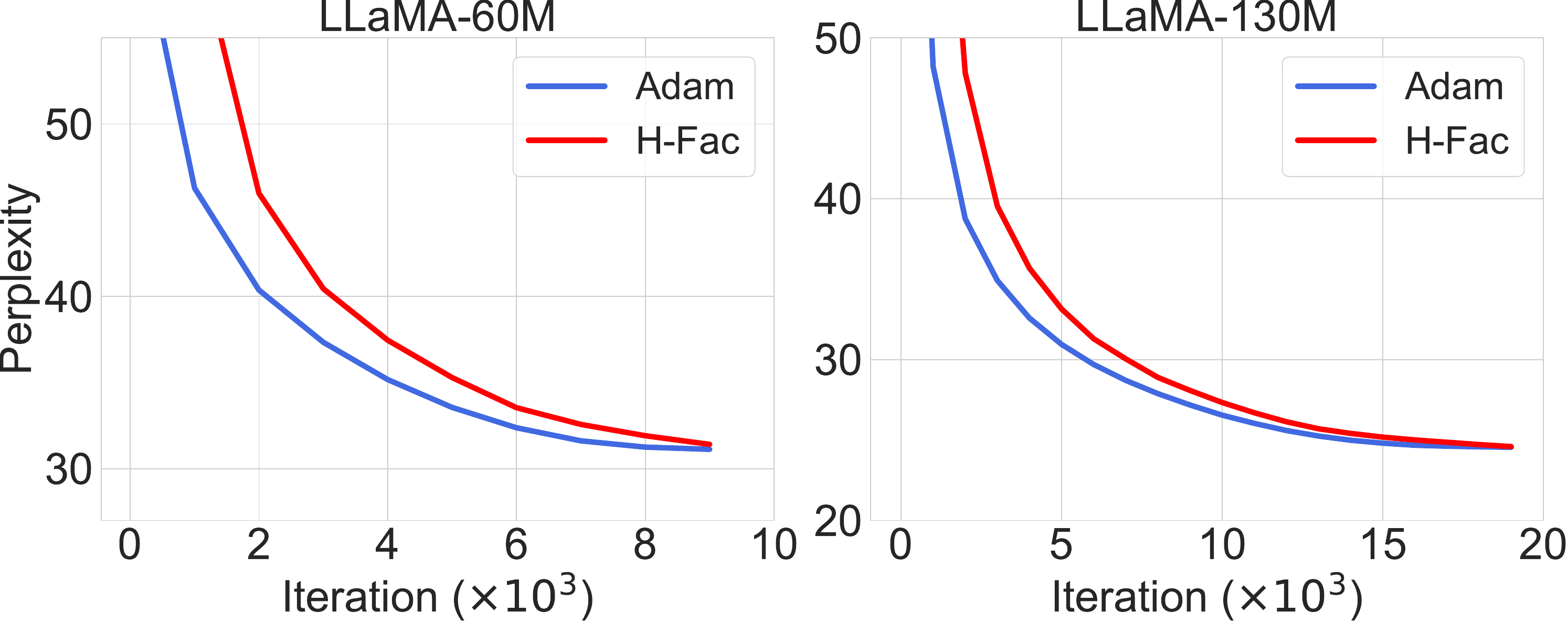}
    \caption{Training progression for pre-training LLaMA models on C4 dataset. Lower is better.}
    \label{fig:llama}
    \vspace{-10pt}
\end{figure}

\paragraph{Evaluation.} As shown in Figure~\ref{fig:llama}, H-Fac demonstrates competitive performances, closely tracking Adam throughout training. At convergence, H-Fac achieves perplexities of 31.41 and 24.59 for the 60M and 130M models, respectively, which are comparable to Adam’s values of 31.12 and 24.55.

\section{CONCLUSION} \label{sec:discussion}
In this study, we represent one of the first attempts to unify and devise adaptive optimization methods through the lens of Hamiltonian dynamics framework. Our approach enables the flexible design of a new class of memory-efficient optimizers, all while preserving theoretical convergence guarantees.
For future works, we propose extending our algorithms to rank-k approximations, aiming to bridge the gap between these full-rank baselines. Another potential direction is to explore optimal projected subspaces, where the row and column means of gradient information can generate better-adapted momentum. For practical use, our current optimizers show promise in applications where ResNets are advantageous because of their efficient training nature. For example, we can adapt our algorithms to the federated learning optimization problem, so that the factored moment estimators can be efficiently communicated to accelerate the convergence. Other applications would also be of interest.

\bibliography{aistast2025}

\begin{thebibliography}{}

\bibitem[Anil et~al., 2019]{anil2019memory}
Anil, R., Gupta, V., Koren, T., and Singer, Y. (2019).
\newblock Memory efficient adaptive optimization.
\newblock {\em Advances in Neural Information Processing Systems}, 32.

\bibitem[Bernstein et~al., 2018]{bernstein2018signsgd}
Bernstein, J., Wang, Y.-X., Azizzadenesheli, K., and Anandkumar, A. (2018).
\newblock signsgd: Compressed optimisation for non-convex problems.
\newblock In {\em International Conference on Machine Learning}, pages 560--569. PMLR.

\bibitem[Bommasani et~al., 2021]{bommasani2021opportunities}
Bommasani, R., Hudson, D.~A., Adeli, E., Altman, R., Arora, S., von Arx, S., Bernstein, M.~S., Bohg, J., Bosselut, A., Brunskill, E., et~al. (2021).
\newblock On the opportunities and risks of foundation models.
\newblock {\em arXiv preprint arXiv:2108.07258}.

\bibitem[Chen et~al., 2023]{chen2023lion}
Chen, L., Liu, B., Liang, K., and Liu, Q. (2023).
\newblock Lion secretly solves constrained optimization: As lyapunov predicts.
\newblock {\em arXiv preprint arXiv:2310.05898}.

\bibitem[Chen et~al., 2024]{chen2024symbolic}
Chen, X., Liang, C., Huang, D., Real, E., Wang, K., Pham, H., Dong, X., Luong, T., Hsieh, C.-J., Lu, Y., et~al. (2024).
\newblock Symbolic discovery of optimization algorithms.
\newblock {\em Advances in Neural Information Processing Systems}, 36.

\bibitem[Cubuk et~al., 2019]{Cubuk2019RandaugmentPA}
Cubuk, E.~D., Zoph, B., Shlens, J., and Le, Q.~V. (2019).
\newblock Randaugment: Practical automated data augmentation with a reduced search space.
\newblock {\em 2020 IEEE/CVF Conference on Computer Vision and Pattern Recognition Workshops (CVPRW)}, pages 3008--3017.

\bibitem[Dettmers et~al., 2021]{dettmers20218}
Dettmers, T., Lewis, M., Shleifer, S., and Zettlemoyer, L. (2021).
\newblock 8-bit optimizers via block-wise quantization.
\newblock {\em arXiv preprint arXiv:2110.02861}.

\bibitem[Dettmers et~al., 2024]{dettmers2024qlora}
Dettmers, T., Pagnoni, A., Holtzman, A., and Zettlemoyer, L. (2024).
\newblock Qlora: Efficient finetuning of quantized llms.
\newblock {\em Advances in Neural Information Processing Systems}, 36.

\bibitem[Devlin et~al., 2018]{devlin2018bert}
Devlin, J., Chang, M.-W., Lee, K., and Toutanova, K. (2018).
\newblock Bert: Pre-training of deep bidirectional transformers for language understanding.
\newblock {\em arXiv preprint arXiv:1810.04805}.

\bibitem[Dosovitskiy et~al., 2020]{dosovitskiy2020image}
Dosovitskiy, A., Beyer, L., Kolesnikov, A., Weissenborn, D., Zhai, X., Unterthiner, T., Dehghani, M., Minderer, M., Heigold, G., Gelly, S., et~al. (2020).
\newblock An image is worth 16x16 words: Transformers for image recognition at scale.
\newblock {\em arXiv preprint arXiv:2010.11929}.

\bibitem[Duchi et~al., 2011]{duchi2011adaptive}
Duchi, J., Hazan, E., and Singer, Y. (2011).
\newblock Adaptive subgradient methods for online learning and stochastic optimization.
\newblock {\em Journal of machine learning research}, 12(7).

\bibitem[Feinberg et~al., 2024]{feinberg2024sketchy}
Feinberg, V., Chen, X., Sun, Y.~J., Anil, R., and Hazan, E. (2024).
\newblock Sketchy: Memory-efficient adaptive regularization with frequent directions.
\newblock {\em Advances in Neural Information Processing Systems}, 36.

\bibitem[Gao et~al., 2022]{gao2022global}
Gao, X., G{\"u}rb{\"u}zbalaban, M., and Zhu, L. (2022).
\newblock Global convergence of stochastic gradient hamiltonian monte carlo for nonconvex stochastic optimization: Nonasymptotic performance bounds and momentum-based acceleration.
\newblock {\em Operations Research}, 70(5):2931--2947.

\bibitem[Gooneratne et~al., 2020]{gooneratne2020low}
Gooneratne, M., Sim, K.~C., Zadrazil, P., Kabel, A., Beaufays, F., and Motta, G. (2020).
\newblock Low-rank gradient approximation for memory-efficient on-device training of deep neural network.
\newblock In {\em ICASSP 2020-2020 IEEE International Conference on Acoustics, Speech and Signal Processing (ICASSP)}, pages 3017--3021. IEEE.

\bibitem[He et~al., 2016]{he2016deep}
He, K., Zhang, X., Ren, S., and Sun, J. (2016).
\newblock Deep residual learning for image recognition.
\newblock In {\em Proceedings of the IEEE conference on computer vision and pattern recognition}, pages 770--778.

\bibitem[Kingma and Ba, 2014]{kingma2014adam}
Kingma, D.~P. and Ba, J. (2014).
\newblock Adam: A method for stochastic optimization.
\newblock {\em arXiv preprint arXiv:1412.6980}.

\bibitem[Kone{\v{c}}n{\`y} et~al., 2016]{konevcny2016federated}
Kone{\v{c}}n{\`y}, J., McMahan, H.~B., Ramage, D., and Richt{\'a}rik, P. (2016).
\newblock Federated optimization: Distributed machine learning for on-device intelligence.
\newblock {\em arXiv preprint arXiv:1610.02527}.

\bibitem[Li et~al., 2024]{li2024memory}
Li, B., Chen, J., and Zhu, J. (2024).
\newblock Memory efficient optimizers with 4-bit states.
\newblock {\em Advances in Neural Information Processing Systems}, 36.

\bibitem[Li et~al., 2020]{li2020federated}
Li, T., Sahu, A.~K., Talwalkar, A., and Smith, V. (2020).
\newblock Federated learning: Challenges, methods, and future directions.
\newblock {\em IEEE signal processing magazine}, 37(3):50--60.

\bibitem[Liu et~al., 2024]{liu2024communication}
Liu, B., Wu, L., Chen, L., Liang, K., Zhu, J., Liang, C., Krishnamoorthi, R., and Liu, Q. (2024).
\newblock Communication efficient distributed training with distributed lion.
\newblock {\em arXiv preprint arXiv:2404.00438}.

\bibitem[Liu et~al., 2023]{liu2023sophia}
Liu, H., Li, Z., Hall, D., Liang, P., and Ma, T. (2023).
\newblock Sophia: A scalable stochastic second-order optimizer for language model pre-training.
\newblock {\em arXiv preprint arXiv:2305.14342}.

\bibitem[Loshchilov and Hutter, 2017]{loshchilov2017decoupled}
Loshchilov, I. and Hutter, F. (2017).
\newblock Decoupled weight decay regularization.
\newblock {\em arXiv preprint arXiv:1711.05101}.

\bibitem[Luo et~al., 2023]{luo2023came}
Luo, Y., Ren, X., Zheng, Z., Jiang, Z., Jiang, X., and You, Y. (2023).
\newblock Came: Confidence-guided adaptive memory efficient optimization.
\newblock {\em arXiv preprint arXiv:2307.02047}.

\bibitem[Maddison et~al., 2018]{maddison2018hamiltonian}
Maddison, C.~J., Paulin, D., Teh, Y.~W., O'Donoghue, B., and Doucet, A. (2018).
\newblock Hamiltonian descent methods.
\newblock {\em arXiv preprint arXiv:1809.05042}.

\bibitem[Mozaffari et~al., 2024]{mozaffari2024mkor}
Mozaffari, M., Li, S., Zhang, Z., and Mehri~Dehnavi, M. (2024).
\newblock Mkor: Momentum-enabled kronecker-factor-based optimizer using rank-1 updates.
\newblock {\em Advances in Neural Information Processing Systems}, 36.

\bibitem[Polyak, 1964]{polyak1964some}
Polyak, B.~T. (1964).
\newblock Some methods of speeding up the convergence of iteration methods.
\newblock {\em Ussr computational mathematics and mathematical physics}, 4(5):1--17.

\bibitem[Raffel et~al., 2020]{raffel2020exploring}
Raffel, C., Shazeer, N., Roberts, A., Lee, K., Narang, S., Matena, M., Zhou, Y., Li, W., and Liu, P.~J. (2020).
\newblock Exploring the limits of transfer learning with a unified text-to-text transformer.
\newblock {\em Journal of machine learning research}, 21(140):1--67.

\bibitem[Reddi et~al., 2019]{reddi2019convergence}
Reddi, S.~J., Kale, S., and Kumar, S. (2019).
\newblock On the convergence of adam and beyond.
\newblock {\em arXiv preprint arXiv:1904.09237}.

\bibitem[Shazeer, 2020]{shazeer2020glu}
Shazeer, N. (2020).
\newblock Glu variants improve transformer.
\newblock {\em arXiv preprint arXiv:2002.05202}.

\bibitem[Shazeer and Stern, 2018]{shazeer2018adafactor}
Shazeer, N. and Stern, M. (2018).
\newblock Adafactor: Adaptive learning rates with sublinear memory cost.
\newblock In {\em International Conference on Machine Learning}, pages 4596--4604. PMLR.

\bibitem[{\c{S}}im{\c{s}}ekli et~al., 2019]{csimcsekli2019heavy}
{\c{S}}im{\c{s}}ekli, U., G{\"u}rb{\"u}zbalaban, M., Nguyen, T.~H., Richard, G., and Sagun, L. (2019).
\newblock On the heavy-tailed theory of stochastic gradient descent for deep neural networks.
\newblock {\em arXiv preprint arXiv:1912.00018}.

\bibitem[Smith et~al., 2021]{smith2021origin}
Smith, S.~L., Dherin, B., Barrett, D.~G., and De, S. (2021).
\newblock On the origin of implicit regularization in stochastic gradient descent.
\newblock {\em arXiv preprint arXiv:2101.12176}.

\bibitem[Szegedy et~al., 2016]{szegedy2016rethinking}
Szegedy, C., Vanhoucke, V., Ioffe, S., Shlens, J., and Wojna, Z. (2016).
\newblock Rethinking the inception architecture for computer vision.
\newblock In {\em Proceedings of the IEEE conference on computer vision and pattern recognition}, pages 2818--2826.

\bibitem[Tian et~al., 2023]{tian2023recent}
Tian, Y., Zhang, Y., and Zhang, H. (2023).
\newblock Recent advances in stochastic gradient descent in deep learning.
\newblock {\em Mathematics}, 11(3):682.

\bibitem[Tieleman, 2012]{tieleman2012lecture}
Tieleman, T. (2012).
\newblock Lecture 6.5-rmsprop: Divide the gradient by a running average of its recent magnitude.
\newblock {\em COURSERA: Neural networks for machine learning}, 4(2):26.

\bibitem[Touvron et~al., 2023]{touvron2023llama}
Touvron, H., Martin, L., Stone, K., Albert, P., Almahairi, A., Babaei, Y., Bashlykov, N., Batra, S., Bhargava, P., Bhosale, S., et~al. (2023).
\newblock Llama 2: Open foundation and fine-tuned chat models.
\newblock {\em arXiv preprint arXiv:2307.09288}.

\bibitem[Vaswani et~al., 2017]{vaswani2017attention}
Vaswani, A., Shazeer, N., Parmar, N., Uszkoreit, J., Jones, L., Gomez, A.~N., Kaiser, {\L}., and Polosukhin, I. (2017).
\newblock Attention is all you need.
\newblock {\em Advances in neural information processing systems}, 30.

\bibitem[Zhang and Sennrich, 2019]{zhang2019root}
Zhang, B. and Sennrich, R. (2019).
\newblock Root mean square layer normalization.
\newblock {\em Advances in Neural Information Processing Systems}, 32.

\bibitem[Zhang et~al., 2017]{Zhang2017mixupBE}
Zhang, H., Ciss{\'e}, M., Dauphin, Y., and Lopez-Paz, D. (2017).
\newblock mixup: Beyond empirical risk minimization.
\newblock {\em ArXiv}, abs/1710.09412.

\bibitem[Zhang et~al., 2020]{zhang2020adaptive}
Zhang, J., Karimireddy, S.~P., Veit, A., Kim, S., Reddi, S., Kumar, S., and Sra, S. (2020).
\newblock Why are adaptive methods good for attention models?
\newblock {\em Advances in Neural Information Processing Systems}, 33:15383--15393.

\bibitem[Zhao et~al., 2024]{zhao2024galore}
Zhao, J., Zhang, Z., Chen, B., Wang, Z., Anandkumar, A., and Tian, Y. (2024).
\newblock Galore: Memory-efficient llm training by gradient low-rank projection.
\newblock {\em arXiv preprint arXiv:2403.03507}.

\bibitem[Zhou et~al., 2020]{zhou2020towards}
Zhou, P., Feng, J., Ma, C., Xiong, C., Hoi, S. C.~H., et~al. (2020).
\newblock Towards theoretically understanding why sgd generalizes better than adam in deep learning.
\newblock {\em Advances in Neural Information Processing Systems}, 33:21285--21296.

\end{thebibliography}
\bibliographystyle{apalike}

\onecolumn

\aistatstitle{Memory-Efficient Optimization with Factorized Hamiltonian Descent: Supplementary Materials}

\setcounter{section}{0}
\renewcommand{\thesection}{\Alph{section}}

\section{HAMILTONIAN FUNCTION ANALYSES} 

\subsection{Factorized momentum-based algorithm} \label{app:first-order}
Recall our general framework for factorized momentum-based optimization methods, which is characterized by the following ODE:
\begin{align*} 
    \dfrac{\mathrm{d}}{\mathrm{d}t} \vv W_t &= 
    -\nabla \phi \big( \beta \hat{\vv u}_t \vv 1_n^\top + \nabla f(\vv W_t) \big)  
    -\nabla \psi \big( \beta \vv 1_m \hat{\vv v}_t^\top + \nabla f(\vv W_t) \big)  \\ 
    \dfrac{\mathrm{d}}{\mathrm{d}t} \vv u_t &= \nabla f(\vv W_t) \vv 1_n / n - \alpha \vv u_t, \quad \quad \hat{\vv u}_t = \vv u_t - \nabla f(\vv W_t) \vv 1_n / n \\    
    \dfrac{\mathrm{d}}{\mathrm{d}t} \vv v_t &= \nabla^\top f(\vv W_t) \vv 1_m / m - \alpha \vv v_t, \quad \quad \hat{\vv v}_t = \vv v_t - \nabla^\top f(\vv W_t) \vv 1_m / m. 
\end{align*}

\noindent
Note that $\vv W_t$ is the weight matrix of size $m \times n$, $\vv u_t$ and $\vv v_t$ are column vectors of size $m$ and $n$, respectively. For a canonical example with $\phi(\vv X) = \psi(\vv X) = \norm{\vv X}_2^2/2$, we have $\dd\phi(\vv X)=\dd\psi(\vv X) = \vv X$, for all matrices $\vv X$. We will show that the following Hamiltonian function: 
$$
\mathcal{H} (\vv W, \vv u, \vv v) = f(\vv W) + \frac{\beta n}{2} \norm{\vv u}_2^2 + \frac{\beta m}{2} \norm{\vv v}_2^2 ,
$$
monotonically decreases along the ODE trajectory above.
Denote $\vv G_t=\nabla f(\vv W_t)$, we can explicitly express the derivative of the model parameter as:
\begin{equation*}
\dfrac{\mathrm{d}}{\mathrm{d}t} \vv W_t = -\left( \beta \vv u_t \vv 1_n^\top + \vv G_t - \frac{\beta}{n} \vv G_t \vv 1_n \vv 1_n^\top \right) - \left( \beta \vv 1_m \vv v_t^\top + \vv G_t - \frac{\beta}{m} \vv 1_m \vv 1_m^\top \vv G_t \right),
\end{equation*}
then we can shorten the derivative of the function $\mathcal{H}(.)$ as:
\begin{align*}
    \frac{\mathrm{d}}{\mathrm{d}t} \mathcal{H}&(\vv W_t, \vv u_t, \vv v_t) \\
    &= \text{trace}\left(\vv G_t^\top \dfrac{\mathrm{d}}{\mathrm{d}t} \vv W_t\right) 
    + \frac{\beta}{n} \vv u_t^\top \dfrac{\mathrm{d}}{\mathrm{d}t} \vv u_t 
    + \frac{\beta}{m} \vv v_t^\top \dfrac{\mathrm{d}}{\mathrm{d}t} \vv v_t \\
    &= - \text{trace} \left(\vv G_t^\top \left(\vv G_t - \frac{\beta}{n} \vv G_t \vv 1_n \vv 1_n^\top \right) \right) 
    - \text{trace} \left(\vv G_t^\top \left(\vv G_t - \frac{\beta}{m} \vv 1_m \vv 1_m^\top \vv G_t \right)\right) 
    - \frac{\alpha \beta}{n} \norm{\vv u_t}_2^2 - \frac{\alpha \beta}{m} \norm{\vv v_t}_2^2 \\
    &= -2 \norm{\vv G_t}_F^2 + \frac{\beta}{n} \norm{\vv G_t \vv 1_n}_2^2 
    + \frac{\beta}{m} \norm{\vv G_t^\top \vv 1_m}_2^2 
    - \frac{\alpha \beta}{n} \norm{\vv u_t}_2^2 
    - \frac{\alpha \beta}{m} \norm{\vv v_t}_2^2,
\end{align*}

in which $\norm{.}_F$ is the Frobenius norm. Since $\beta \in (0,1)$, applying C-S inequality, we have:
\begin{equation}
\frac{\mathrm{d}}{\mathrm{d}t} \mathcal{H}(\vv W_t, \vv u_t, \vv v_t) \leq -2 \norm{\vv G_t}_F^2 + \frac{1}{n} \norm{\vv G_t \vv 1_n}_2^2 + \frac{1}{m} \norm{\vv G_t^\top \vv 1_m}_2^2 \leq 0. \tag*{\qed}
\end{equation}

\subsection{Adafactor} \label{app:adafactor}
Recall the ODE of Adafactor with first-order moment integration:
\begin{align*}
    \dfrac{\mathrm{d}}{\mathrm{d}t} \vv W_t &= -\frac{\vv M_t}{\sqrt{\vv r_t \vv s_t^\top / \vv 1_m^\top \vv r_t}}, \quad \quad \dfrac{\mathrm{d}}{\mathrm{d}t} \vv M_t = \nabla f(\vv W_t) - \alpha \vv M_t \\
    \dfrac{\mathrm{d}}{\mathrm{d}t} \vv r_t &= (\nabla f(\vv W_t))^2 \vv 1_n - \beta \vv r_t, \quad \quad \dfrac{\mathrm{d}}{\mathrm{d}t} \vv s_t = (\nabla^\top f(\vv W_t))^2 \vv 1_m - \beta \vv s_t 
\end{align*}
\newpage
We will show that the Hamiltonian function is described by:
$$
\mathcal{H}(\vv W, \vv M, \vv r, \vv s) = f(\vv W) + \underbrace{\frac{1}{2} \sum_{i=1, j=1}^{m, n} \frac{M_{ij}^2 \sqrt{\sum_{i=1}^m r_i}}{\sqrt{r_is_j}}}_{\mathcal{R}(\vv M, \vv r, \vv s)},
$$ 
is monotonically decreasing along the ODE trajectory when $\beta \leq 2 \alpha$, in which $M_{ij}$ is the element $(i, j)$-th of the first-moment matrix $\vv M$, $r_i$ and $s_j$ represent the $i$-th and $j$-th elements of the column vectors $\vv r$ and $\vv s$, respectively.
\noindent
Denote \( \vv G_t = \nabla f(\vv W_t) \) and \( \mathcal{R}_t = \mathcal{R}(\vv M_t, \vv r_t, \vv s_t) \), then the derivative of \( \mathcal{H}(.) \) can be expressed as:
$$
\frac{\mathrm{d}}{\mathrm{d}t} \mathcal{H}(\vv W_t, \vv M_t, \vv r_t, \vv s_t) = 
    \text{trace}\left(\vv G_t^\top \frac{\mathrm{d}}{\mathrm{d}t} \vv W_t \right) 
    + \text{trace}\left((\nabla_{\vv M} \mathcal{R}_t)^\top \frac{\mathrm{d}}{\mathrm{d}t} \vv M_t \right) 
    + (\nabla_{\vv r} \mathcal{R}_t)^\top \frac{\mathrm{d}}{\mathrm{d}t} \vv r_t 
    + (\nabla_{\vv s} \mathcal{R}_t)^\top \frac{\mathrm{d}}{\mathrm{d}t} \vv s_t.
$$
Analyzing each element specifically, we get:
\begin{align*}
    \text{trace}(\vv G_t^\top \frac{\mathrm{d}}{\mathrm{d}t} \vv W_t) &= -  \text{trace}\left(\vv G_t^\top \frac{\vv M_t \sqrt{\vv 1_m^\top \vv r_t}}{\sqrt{\vv r_t \vv s_t^\top }}\right),  \\
    \text{trace}\left((\nabla_{\vv M} \mathcal{R}_t)^\top \frac{\mathrm{d}}{\mathrm{d}t} \vv M_t \right) 
    &= \text{trace}\left(\left(\vv G_t - \alpha \vv M_t\right)^\top 
    \frac{\vv M_t \sqrt{\vv 1_m^\top \vv r_t}}{\sqrt{\vv r_t \vv s_t^\top }}\right) = \text{trace}\left(\vv G_t^\top 
    \frac{\vv M_t \sqrt{\vv 1_m^\top \vv r_t}}{\sqrt{\vv r_t \vv s_t^\top }}\right) 
    - \alpha \, \text{trace}\left(\vv M_t^\top 
    \frac{\vv M_t \sqrt{\vv 1_m^\top \vv r_t}}{\sqrt{\vv r_t \vv s_t^\top }}\right),
\end{align*}

\begin{align*}
    (\nabla_{\vv r} \mathcal{R}_t)^\top \frac{\mathrm{d}}{\mathrm{d}t} \vv r_t &= \sum_{i=1}^m (\partial_{r_i} \mathcal{R}_t)(\vv 1_n^\top \vv G_{t,i}^2 - \beta r_{t,i})  \tag*{\textcolor{magenta}{// $\vv G_{t,i}$ is the $i$-th row of $\vv G_t$, and $r_{t,i}$ is the $i$-th element of $\vv r_t$}} \\
    &= \sum_{i=1}^m \left(\frac{1}{2} \sum_{j=1}^n \frac{M_{t, ij}^2}{\sqrt{s_{t, j}}} \right) \frac{-\sum_{k \neq i} r_{t,k}}{2 \sqrt{r_{t,i}^{3/2}} \sqrt{\sum_i r_{t,i}}}(\vv 1_n^\top \vv G_{t,i}^2 - \beta r_{t,i}) \tag*{\textcolor{magenta}{// $M_{t,ij}$ is the $(i,j)$-th element of $\vv M_t$, and $s_{t,j}$ is the $j$-th element of $\vv s_t$}}\\
    &\leq \sum_{i=1}^m \left(\frac{\beta}{4} \sum_{j=1}^n \frac{M_{t,ij}^2}{\sqrt{s_{t,j}}} \right) \frac{\sum_{k \neq i} r_{t,k}}{\sqrt{r_{t,i}} \sqrt{\sum_i r_{t,i}}} \tag*{\textcolor{magenta}{// the multiplication by $\vv 1_n^\top \vv G_{t,i}^2$ is $\leq 0$}} \\
    &\leq \sum_{i=1}^m \left(\frac{\beta}{4} \sum_{j=1}^n \frac{M_{t,ij}^2}{\sqrt{s_{t,j}}} \right) \frac{\sqrt{\sum_{i} r_{t,i}}}{\sqrt{r_{t,i}} } \tag*{\textcolor{magenta}{// $\sum_{k \neq i} r_{t,k} \leq \sum_i r_{t,i}$}} \\
    &= \sum_{i=1}^m \left(\frac{\beta}{4} \sum_{j=1}^n \frac{M_{t,ij}^2 \sqrt{\sum_{i} r_{t,i}}}{\sqrt{r_{t,i} s_{t,j}}} \right)  \\
    &=  \frac{\beta}{4} \text{trace}\left(\vv M_t^\top \frac{\vv M_t \sqrt{\vv 1_m^\top \vv r_t}}{\sqrt{\vv r_t \vv s_t^\top }}\right), \\
    (\nabla_{\vv s} \mathcal{R}_t)^\top \frac{\mathrm{d}}{\mathrm{d}t} \vv s_t &=  \sum_{j=1}^n (\partial_{s_j} \mathcal{R}_t)(\vv 1_m^\top \vv G_{t, :j}^2 - \beta s_{t,j}) \tag*{\textcolor{magenta}{// $\vv G_{t, :j}$ is the $j$-th column of $\vv G_t$}} \\
    &= \sum_{j=1}^n \left(\frac{1}{2} \sum_{i=1}^m \frac{M_{t,ij}^2 \sqrt{\sum_i r_{t,i}}}{\sqrt{r_{t,i}}} \right) \frac{-1}{2 \sqrt{s_{t,j}^{3/2}}} (\vv 1_m^\top \vv G_{t,:j}^2 - \beta s_{t,j}) \\
    &\leq \sum_{j=1}^n \left(\frac{\beta}{4} \sum_{i=1}^m \frac{M_{t,ij}^2 \sqrt{\sum_i r_{t,i}}}{\sqrt{r_{t,i} s_{t,j}}} \right) \tag*{\textcolor{magenta}{// the multiplication by $\vv 1_m^\top \vv G_{t,:j}^2$ is $\leq 0$}}\\
    &=\frac{\beta}{4} \text{trace}\left(\vv M_t^\top \frac{\vv M_t \sqrt{\vv 1_m^\top \vv r_t}}{\sqrt{\vv r_t \vv s_t^\top }}\right). 
\end{align*}
Canceling out similar quantities, we can rewrite the Hamiltonian derivative as follows:

\begin{equation}
\frac{\mathrm{d}}{\mathrm{d}t} \mathcal{H}(\vv W_t, \vv M_t, \vv r_t, \vv s_t) \leq \left( \frac{\beta}{2} - \alpha \right) \text{trace}\left(\vv M_t^\top \frac{\vv M_t \sqrt{\vv 1_m^\top \vv r_t}}{\sqrt{\vv r_t \vv s_t^\top }}\right) = \left( \frac{\beta}{2} - \alpha \right) \sum_{i=1}^m \sum_{j=1}^n \frac{M_{t, ij}^2 \sqrt{\sum_i r_{t,i}}}{\sqrt{r_{t,i}s_{t,j}}}  \leq 0. \tag*{\qed} 
\end{equation}


\subsection{H-Fac} \label{app:hfac}

In this part, we will prove that the following ODE trajectory (with $\beta \leq 4\alpha$):
\begin{align*}
    \frac{\mathrm{d}}{\mathrm{d}t} \vv W_t &= 
    - \frac{1}{2} \bigg(
    \frac{\vv u_t \vv 1_n^\top - \nabla f(\vv W_t) \vv 1_n \vv 1_n^\top / n}{\sqrt{\vv r_t \vv 1_n^\top}} 
    + \frac{\vv 1_m \vv v_t^\top - \vv 1_m \vv 1_m^\top \nabla f(\vv W_t) / m}{\sqrt{\vv 1_m \vv s_t^\top}}
    \bigg) 
    - \frac{\nabla f(\vv W_t)}{\sqrt{\vv r_t \vv s_t^\top / \vv 1_m^\top \vv r_t}} \\
    \frac{\mathrm{d}}{\mathrm{d}t} \vv u_t &= \nabla f(\vv W_t) \vv 1_n / n - \alpha \vv u_t \\
    \frac{\mathrm{d}}{\mathrm{d}t} \vv v_t &= \nabla^\top f(\vv W_t) \vv 1_m / m - \alpha \vv v_t \\
    \frac{\mathrm{d}}{\mathrm{d}t} \vv r_t &= (\nabla f(\vv W_t))^2 \vv 1_n - \beta \vv r_t \\
    \frac{\mathrm{d}}{\mathrm{d}t} \vv s_t &= (\nabla^\top f(\vv W_t))^2 \vv 1_m - \beta \vv s_t
\end{align*}
descends the Hamiltonian function defined by:
$$
\mathcal{H}(\vv W, \vv u, \vv v, \vv r, \vv s) := f(\vv W) + \underbrace{\frac{n}{4} \sum_{i=1}^m \frac{u_i^2}{\sqrt{r_i}} + \frac{m}{4} \sum_{j=1}^n \frac{v_j^2}{\sqrt{s_j}}}_{\mathcal{R}(\vv u, \vv v, \vv r, \vv s)}.
$$
with $u_i, r_i$ represent the $i$-th elements of the column vectors $\vv u, \vv r$, and $v_j, s_j$ represent the $j$-th elements of the column vectors $\vv v, \vv s$.
Denote \( \vv G_t = \nabla f(\vv W_t) \) and \( \mathcal{R}_t = \mathcal{R}(\vv u_t, \vv v_t, \vv r_t, \vv s_t) \), we have:
$$
\frac{\mathrm{d}}{\mathrm{d}t} \mathcal{H}(\vv W_t, \vv u_t, \vv v_t, \vv r_t, \vv s_t) = 
\text{trace}\left(\vv G_t^\top \frac{\mathrm{d}}{\mathrm{d}t} \vv W_t\right) 
+ (\nabla_{\vv u} \mathcal{R}_t)^\top \frac{\mathrm{d}}{\mathrm{d}t} \vv u_t 
+ (\nabla_{\vv v} \mathcal{R}_t)^\top \frac{\mathrm{d}}{\mathrm{d}t} \vv v_t 
+ (\nabla_{\vv r} \mathcal{R}_t)^\top \frac{\mathrm{d}}{\mathrm{d}t} \vv r_t 
+ (\nabla_{\vv s} \mathcal{R}_t)^\top \frac{\mathrm{d}}{\mathrm{d}t} \vv s_t.
$$

Similar to the previous part, we can calculate each element specifically as follows:
\begin{align*}
    \text{trace}(\vv G_t^\top \frac{\mathrm{d}}{\mathrm{d}t} \vv W_t) 
    &= - \frac{1}{2} \text{trace}\left(\vv G_t^\top \frac{\vv u_t \vv 1_n^\top}{\sqrt{\vv r_t \vv 1_n^\top}}\right) 
    - \frac{1}{2} \text{trace}\left(\vv G_t^\top \frac{\vv 1_m \vv v_t^\top}{\sqrt{\vv 1_m \vv s_t^\top}}\right) \\
    &\quad + \text{trace}\left(\vv G_t^\top \frac{\vv G_t \vv 1_n \vv 1_n^\top}{2n \sqrt{\vv r_t \vv 1_n^\top}} \right) 
    + \text{trace}\left(\vv G_t^\top \frac{\vv 1_m \vv 1_m^\top \vv G_t}{2m \sqrt{\vv 1_m \vv s_t^\top}} \right) 
    - \text{trace}\left(\vv G_t^\top \frac{\vv G_t \sqrt{\vv 1_m^\top \vv r_t}}{\sqrt{\vv r_t \vv s_t^\top}} \right), \\
    (\nabla_{\vv u} \mathcal{R}_t)^\top \frac{\mathrm{d}}{\mathrm{d}t} \vv u_t 
    &= \frac{n}{2} \left( \frac{\vv u_t}{\sqrt{\vv r_t}} \right)^\top \left(\vv G_t \vv 1_n / n - \alpha \vv u_t\right) 
    = \frac{1}{2} \left( \frac{\vv u_t}{\sqrt{\vv r_t}} \right)^\top \vv G_t \vv 1_n - \frac{n \alpha}{2} \left( \frac{\vv u_t}{\sqrt{\vv r_t}} \right)^\top \vv u_t, \\
    (\nabla_{\vv v} \mathcal{R}_t)^\top \frac{\mathrm{d}}{\mathrm{d}t} \vv v_t 
    &= \frac{m}{2} \left( \frac{\vv v_t}{\sqrt{\vv s_t}} \right)^\top \left( \vv G_t^\top \vv 1_m / m - \alpha \vv v_t\right) 
    = \frac{1}{2} \left( \frac{\vv v_t}{\sqrt{\vv s_t}} \right)^\top \vv G_t^\top \vv 1_m - \frac{m \alpha}{2} \left( \frac{\vv v_t}{\sqrt{\vv s_t}} \right)^\top \vv v_t, \\
    (\nabla_{\vv r} \mathcal{R}_t)^\top \frac{\mathrm{d}}{\mathrm{d}t} \vv r_t 
    &= \left( -\frac{n}{8} \frac{\vv u_t^2}{\sqrt{\vv r_t^{3/2}}} \right)^\top  \left(\vv G_t^2 \vv 1_n - \beta \vv r_t\right) 
    \leq \frac{n\beta}{8} \left(\frac{\vv u_t^2}{\sqrt{\vv r_t^{3/2}}} \right)^\top \vv r_t, 
    \tag*{\textcolor{magenta}{// the multiplication by $\vv G_t^2 \vv 1_n$ is $\leq 0$}} \\
    (\nabla_{\vv s} \mathcal{R}_t)^\top \frac{\mathrm{d}}{\mathrm{d}t} \vv s_t 
    &= \left( -\frac{m}{8} \frac{\vv v_t^2}{\sqrt{\vv s_t^{3/2}}} \right)^\top  \left(\left(\vv G_t^2\right)^\top \vv 1_m - \beta \vv s_t\right) 
    \leq \frac{m \beta}{8} \left( \frac{\vv v_t^2}{\sqrt{\vv s_t^{3/2}}} \right)^\top \vv s_t. 
    \tag*{\textcolor{magenta}{// the multiplication by $(\vv G_t^2)^\top \vv 1_m$ is $\leq 0$}}
\end{align*}

\noindent
Canceling out crossing terms:
\begin{align*}
    \text{trace}\left(\vv G_t^\top \frac{\vv u_t \vv 1_n^\top}{\sqrt{\vv r_t \vv 1_n^\top}}\right) 
    &= \left( \frac{\vv u_t}{\sqrt{\vv r_t}} \right)^\top \vv G_t \vv 1_n, 
    \quad \quad \text{trace} \left(\vv G_t^\top \frac{\vv 1_m \vv v_t^\top}{\sqrt{\vv 1_m \vv s_t^\top}}\right) 
    = \left( \frac{\vv v_t}{\sqrt{\vv s_t}} \right)^\top \vv G_t^\top \vv 1_m \\ 
    \left( \frac{\vv u_t}{\sqrt{\vv r_t}} \right)^\top \vv u_t &= \left(\frac{\vv u_t^2}{\sqrt{\vv r_t^{3/2}}} \right)^\top \vv r_t , 
    \quad \quad \left( \frac{\vv v_t}{\sqrt{\vv s_t}} \right)^\top \vv v_t = \left( \frac{\vv v_t^2}{\sqrt{\vv s_t^{3/2}}} \right)^\top \vv s_t
\end{align*}
and with a note that $\beta \leq 4\alpha$, we can rewrite the Hamiltonian derivative as follows:
\begin{align*}
\frac{\mathrm{d}}{\mathrm{d}t} \mathcal{H}_t  &\leq 
\text{trace}\left(\vv G_t^\top \frac{\vv G_t \vv 1_n \vv 1_n^\top}{2n \sqrt{\vv r_t \vv 1_n^\top}} \right) 
+ \text{trace}\left(\vv G_t^\top \frac{\vv 1_m \vv 1_m^\top \vv G_t}{2m \sqrt{\vv 1_m \vv s_t^\top}} \right) 
- \text{trace} \left(\vv G_t^\top \frac{\vv G_t \sqrt{\vv 1_m^\top \vv r_t}}{\sqrt{\vv r_t \vv s_t^\top}} \right) \\
&= \text{trace}\left( \frac{1}{n} (\vv G_t \vv 1_n)^2 \frac{1}{2 \sqrt{\vv r_t^\top}} \right) 
+ \text{trace}\left( \frac{1}{m} (\vv G_t^\top \vv 1_m)^2 \frac{1}{2 \sqrt{\vv s_t^\top}} \right) 
- \text{trace} \left( \vv G_t^2 \frac{\sqrt{\vv 1_m^\top \vv r_t}}{\sqrt{\vv s_t \vv r_t^\top}} \right).
\end{align*}
Applying C-S inequality gives us \( \dfrac{1}{n} (\vv G_t \vv 1_n)^2 \leq \vv G_t^2 \vv 1_n \) and \( \dfrac{1}{m} (\vv G_t^\top \vv 1_m)^2 \leq (\vv G_t^\top)^2 \vv 1_m \), therefore: 
\begin{align*}
\frac{\mathrm{d}}{\mathrm{d}t} \mathcal{H}_t &\leq 
\text{trace}\left( \vv G_t^2 \vv 1_n \frac{1}{2 \sqrt{\vv r_t^\top}} \right) 
+ \text{trace}\left( (\vv G_t^\top)^2 \vv 1_m \frac{1}{2 \sqrt{\vv s_t^\top}} \right) 
- \text{trace} \left( \vv G_t^2 \frac{\sqrt{\vv 1_m^\top \vv r_t}}{\sqrt{\vv s_t \vv r_t^\top}} \right) \\
&= \frac{1}{2} \text{trace} \bigg[ \vv G_t^2 \bigg( 
\underbrace{\vv 1_n \frac{1}{\sqrt{\vv r_t^\top}} + \frac{1}{\sqrt{\vv s_t}} \vv 1_m^\top 
- 2 \frac{\sqrt{\vv 1_m^\top \vv r_t}}{\sqrt{\vv s_t \vv r_t^\top}}}_{\mathcal{F}} 
\bigg) \bigg].
\end{align*}
By zero initialization, we note that the moving averages of row sums and column sums are symmetric, in other words, $\vv 1_m^\top \vv r_t$ and $\vv 1_n^\top \vv s_t$ both represent the moving average of the sum of all squared gradient entries. Denote this quantity by $\mathcal{S}$, we have $\mathcal{S} = \vv 1_m^\top \vv r_t = \vv 1_n^\top \vv s_t \geq (r_{t,i} + s_{t,j})/2$ for all $i, j$, then consider each element of matrix $\mathcal{F}$:
$$
\frac{1}{\sqrt{r_{t,i}}} + \frac{1}{\sqrt{s_{t,j}}} - 2\frac{\sqrt{\mathcal{S}}}{\sqrt{r_{t,i}s_{t,j}}} \leq \frac{1}{\sqrt{r_{t,i}}} + \frac{1}{\sqrt{s_{t,j}}} - \frac{\sqrt{2(r_{t,i} + s_{t,j})}}{\sqrt{r_{t,i} s_{t,j}}} \leq 0 ,
$$
by C-S inequality. Therefore $\mathcal{F}$ is a negative matrix, and as a result, $\dfrac{\mathrm{d}}{\mathrm{d}t} \mathcal{H}(\vv W_t, \vv u_t, \vv v_t, \vv r_t, \vv s_t) \leq 0$. \qed

\paragraph{Convergence to local optimal.}The above optimization algorithms all ensure that their Hamiltonian $\mathcal{H}(\vv W_t, \vv S_t)$ are monotonically decreasing along the corresponding ODE trajectories, where $\vv S_t$ represents the optimization states in general. Consequently, by applying LaSalle’s Invariance principle, the set of accumulation points $(\vv W_t, \vv S_t)$ must lie within the set $\mathcal{I}$, where $\mathcal{I} = \{\text{the union of complete trajectories satisfying }$ $\dfrac{\mathrm{d}}{\mathrm{d}t} \mathcal{H}(\vv W_t, \vv S_t) = 0, \forall t \}$. Based on the preceding inequality transformations, it is evident that the points in the limit set $\mathcal{I}$ must satisfy $\nabla f(\vv W_t) \equiv 0$. This implies that those trajectories will converge to local optima.


\section{FACTORIZED LION OPTIMIZER} \label{app:lionfactor}
Recently, a new optimization named Lion (Evolved Sign Momentum)~\citep{chen2024symbolic} was discovered by an evolutionary search algorithm applied to a symbolically represented program space. Lion has been shown to achieve at least comparable performance to AdamW on a wide range of tasks while reducing memory cost and training time. Notably, Lion can be formulated as an iterative update procedure:
\begin{align*}
    \vv M_{t} &= \beta_2 \vv M_{t-1} + (1 - \beta_2) \nabla f(\vv W_{t}) \\
    \vv W_{t+1} &= \vv W_{t} - \eta_t \left( 
    \text{sign}\left(\beta_1 \vv M_{t-1} + (1 - \beta_1) \nabla f(\vv W_{t})\right) 
    + \lambda \vv W_{t} \right) \tag*{\textcolor{magenta}{// Lion}}
\end{align*}

\begin{minipage}[t]{0.45\textwidth}
\vspace{0.33in}
\begin{algorithm}[H]
\caption{\textcolor{magenta}{Lion}.}
\label{alg:lion}
\begin{algorithmic}
\STATE \textbf{Inputs:} \textit{double}-moment coefficients $\beta_1=0.9$, $\beta_2=0.99$, and regularization constant $\lambda$
\STATE \textbf{Initialization:} weight parameters $\vv W_1 \in \mathbb{R}^{m \times n}$, initial momentum $\vv M_0 \leftarrow 0$
\FOR{$t=1$ to $T$}
    \STATE $\vv G_t = \nabla f_t(\vv W_{t})$
    \STATE \textbf{update model parrameters}
    \STATE $\vv C_{t} = \beta_1 \vv M_{t-1} + (1-\beta_1) \vv G_t $
    \STATE $\vv W_{t+1} = \vv W_{t} - \eta_{t} \left( \sign (\vv C_t) + \lambda \vv W_{t} \right)$
    \STATE \textbf{update exponential moment averages}
    \STATE $\vv M_{t} = \beta_2 \vv M_{t-1} + (1-\beta_2) \vv G_t $
\ENDFOR
\end{algorithmic}
\end{algorithm}
\end{minipage}
\hspace{0.5cm}
\begin{minipage}[t]{0.50\textwidth}
\begin{algorithm}[H]
\caption{\textcolor{magenta}{Lionfactor} for matrix parameter, with factored first-order moments.}
\label{alg:lionfactor}
\begin{algorithmic}
\STATE \textbf{Inputs:} \textit{double}-moment coefficients $\beta_1 = 0.9$, $\beta_2 = 0.99$, and regularization constant $\lambda$
\STATE \textbf{Initialization:} weight parameters $\vv W_1 \in \mathbb{R}^{m \times n}$, initial moment factors $\vv u_0, \vv v_0 \leftarrow 0$
\FOR{$t = 1$ to $T$}
    \STATE $\vv G_t = \nabla f_t(\vv W_{t})$
    \STATE \textbf{Update model parameters:}
    \STATE $\hat{\vv u}_{t} = \beta_1 \vv u_{t-1} + (1 - \beta_1) \vv G_t \vv 1_n / n - \vv G_t \vv 1_n / n$
    \STATE $\hat{\vv v}_{t} = \beta_1 \vv v_{t-1} + (1 - \beta_1) \vv G_t^\top \vv 1_m / m - \vv G_t^\top \vv 1_m / m$
    \STATE $\vv W_{t+1} = \vv W_{t} - \eta_t \left( \sign \left(\hat{\vv u}_{t} \vv 1_n^\top + \vv G_t \right) \right.$ 
    \STATE \hspace{3cm} $ +\ \sign (\vv 1_m \hat{\vv v}_{t}^\top + \vv G_t) + \lambda \vv W_{t})$
    \STATE \textbf{Update exponential moment averages:}
    \STATE $\vv u_{t} = \beta_2 \vv u_{t-1} + (1 - \beta_2) \vv G_t \vv 1_n / n$
    \STATE $\vv v_{t} = \beta_2 \vv v_{t-1} + (1 - \beta_2) \vv G_t^\top \vv 1_m / m$
\ENDFOR
\end{algorithmic}
\end{algorithm}
\vspace{0.1in}
\end{minipage}

We can see that when $\beta_2 = \beta_1$, Lion will resemble \textit{sign}SGD with momentum~\citep{bernstein2018signsgd}. However, Lion used a double$-\beta$ scheme with default values $\beta_1=0.9, \beta_2=0.99$. Intuitively, this allows Lion to remember longer the gradient history accumulated by the momentum, meanwhile assign a higher weight to the current gradient. Comprehensive experimental results show that Lion converges faster and usually generalizes better than AdamW, but with greater memory efficiency as it only keeps track of the momentum. 

\vspace{1cm}

\begin{figure}[h!]
    \centering
    \includegraphics[width=\textwidth]{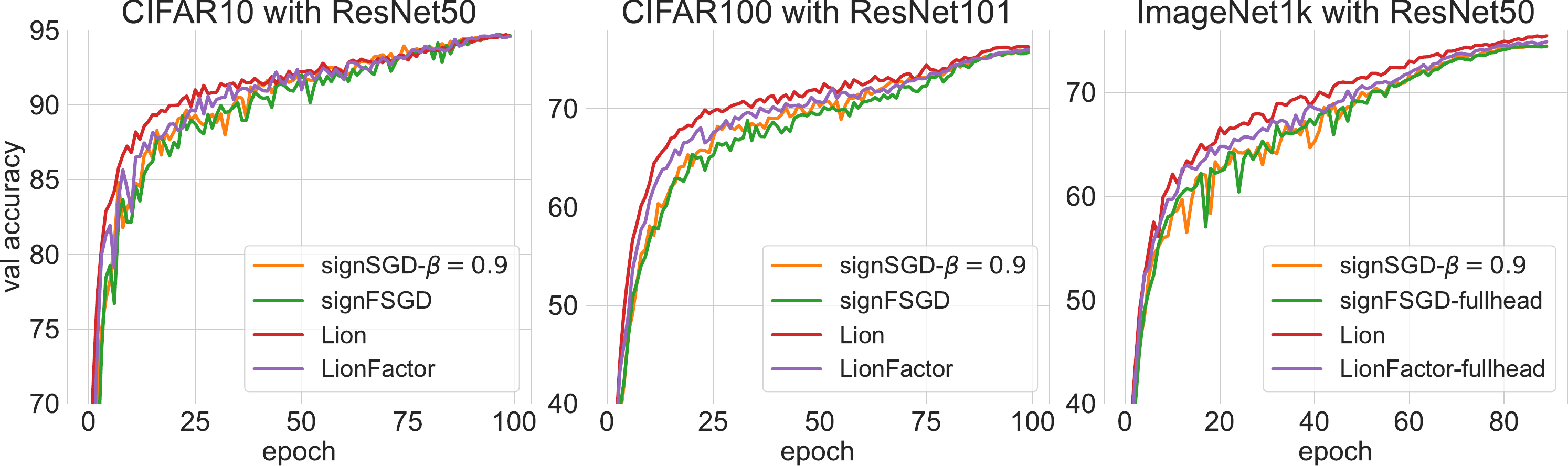}
    \caption{Performance of sign-based optimizers on ResNet architectures. ``fullhead'' means the version using full momentum for the MLP head layer.}
    \label{fig:lionfactor}
\end{figure}

\vspace{1cm}

We can similarly apply the double$-\beta$ scheme to our \textit{sign}FGSD, and obtain a new algorithm that we call \textcolor{magenta}{LionFactor}~\ref{alg:lionfactor}. We conducted several experiments to evaluate the performance of LionFactor on ResNet models. The results are shown in Figure~\ref{fig:lionfactor}. Interestingly, LionFactor performs significantly better than \textit{sign}FSGD, even \textit{sign}SGD with momentum, in terms of both convergence rate and accuracy. Although LionFactor still shares the same drawbacks with \textit{sign}FSGD when applied to models such as ViTs, it makes a lot of sense to explore more efficient algorithms to factorize the momentum in Lion optimizer. We leave it for future work.

\newpage

\section{NUMERICAL RESULTS}
\vspace{-0.1in}
\begin{table*}[h!]
    \centering
    \caption{ImageNet1K top-1 accuracy results for various optimizers on different models.}
    \begin{tabular}{l cc c c c cc c}
        \toprule
         \textbf{Models} & \multicolumn{2}{c}{\textbf{\textit{sign}SGD}} & \textbf{\textit{sign}FSGD} & \textbf{\textit{sign}FSGD} & \textbf{AdamW} & \multicolumn{2}{c}{\textbf{Adafactor}} & \textbf{H-Fac} \\ 
         \cmidrule{2-3} \cmidrule{5-5} \cmidrule{7-8}
        & $m=0.9$ & $m=0.0$ & & \textit{fullhead} & & $m=0.9$ & $m=0.0$ &  \\ 
        \midrule
        ResNet50 & 74.47 & 72.36 & 72.47 & 74.45 & 75.61 & 75.85 & 75.79 & \textbf{75.90} \\
        ViT-B/32 & \multicolumn{2}{c}{} & \multicolumn{2}{c}{} & 72.20 & \textbf{72.31} & 71.36 & 71.87 \\
        ViT-S/16 & \multicolumn{2}{c}{} & \multicolumn{2}{c}{} & \textbf{78.35} & 77.81 & 76.74 & 77.20 \\ 
        \bottomrule
    \end{tabular}
    \label{tab:imagenet}
\end{table*}

\begin{table}[h!]
    \centering
    \caption{Memory requirements for different optimizers, with weight parameter of size $m \times n$}
    \begin{tabular}{lcccccc}
        \toprule
         & \textbf{\textit{sign}SGD} & {\textbf{\textit{sign}FSGD}} & \textbf{Adam} & \multicolumn{2}{c}{\textbf{Adafactor}} & {\textbf{H-Fac}} \\
         \cmidrule{2-2} \cmidrule{5-6}
        & \textit{w/ momentum} & & &  \textit{w/ momentum} & \textit{w/o momentum}  \\
        \midrule
        Weights & $mn$ & $mn$ & $mn$ & $mn$ & $mn$ & $mn$   \\
        Gradient & $mn$ & $mn$ & $mn$ & $mn$ & $mn$ & $mn$   \\
        Optim. States & $mn$ & $m+n$ & $2mn$ & $mn+m+n$ & $m+n$ & $2(m+n)$  \\
        \bottomrule
    \end{tabular}
    \label{tab:optimizer_memory}
\end{table}
We provide in Table~\ref{tab:optimizer_memory} the memory cost of optimizers implemented in this paper. Our proposed optimizers, \textit{sign}FSGD and H-Fac,  offer sublinear memory costs, which are comparable to those of conventional gradient descent (without momentum). In our implementation, the rank-1 parameterization enables in-place vector operators that do not require additional memory.

\section{HYPERPARAMETER SETTINGS}\label{app:hyper-settings}
We conducted experiments in distributed setup on 8 V100 GPUs, using Accelerate library from Hugging Face.

For the image classification task, we opted for recommended configurations of learning rate ($lr$), weight decay ($\lambda$), and dropout rate ($dr$) from prior research. The detailed settings are given in Table~\ref{tab:img_settings}.
\begin{table}[h!]
    \centering
    \caption{Hyperparameters for the experiments of pre-training ResNet50 and ViTs on ImagNet1k}
    \begin{tabular}{lcccccccccc}
        \toprule
        \multirow{2}{*}{\textbf{Optimizers}} & \multirow{2}{*}{$\beta_1$} & \multirow{2}{*}{$\beta_2$} & \multirow{2}{*}{$\epsilon$} & \multicolumn{2}{c}{\textbf{ResNet50}} & \multicolumn{5}{c}{\textbf{ViT-B/32, ViT-S/16}}  \\
        \cmidrule(lr){5-6} \cmidrule(lr){7-11} 
        & & & & $lr$ & $\lambda$ & $lr$ & $\lambda$ & $dr$ & RandAug & Mixup  \\
        \midrule
        \textit{sign}SGD & 0.9, 0.0 & - & - & 0.0003 & 1.0 & - & - & - & - & - \\
        \textit{sign}FSGD  & 0.9 & - & - & 0.0003 & 1.0 & - & - & - & - & - \\
        Lion & 0.9 & 0.99 & - & 0.0003 & 1.0 & - & - & - & - & - \\
        Lionfactor & 0.9 & 0.99 & - & 0.0003 & 1.0 & - & - & - & - & - \\
        \midrule
        Adam & 0.9 & 0.999 & 1e-8 & 0.001 & 0.1 & 0.003 & 0.1 & 0.0 & 2, 15 & 0.5\\
        Adafactor & 0.9, 0.0 & 0.999 & 1e-30 & 0.001 & 0.1 & 0.003 & 0.1 & 0.0 & 2, 15 & 0.5 \\ 
        H-Fac & 0.9 & 0.999 & 1e-30 & 0.001 & 0.1 & 0.003 & 0.1 & 0.0 & 2, 15 & 0.5 \\
        \bottomrule
    \end{tabular}
    \label{tab:img_settings}
\end{table}

For language modeling, we tuned the learning rate over the set \{0.003, 0.001, 0.0003, 0.0001\} and selected the optimal value based on validation perplexity. The specific settings are summarized in the Table~\ref{tab:llama_settings}.
\begin{table}[h!]
    \centering
    \caption{Training configuration for LLaMA models}
    \begin{tabular}{lcccc}
        \toprule
        \textbf{LLaMA models} & \textbf{Tokens} & \textbf{Training Steps} & \textbf{Warmup Steps} & \textbf{Learning Rate} \\
        \midrule
        60M & 1.3B & 10K & 1K & 0.003 \\
        130M & 2.6B & 20K & 2K & 0.001 \\
        \bottomrule
    \end{tabular}
    \label{tab:llama_settings}
\end{table}

\end{document}